\documentclass{article}
\usepackage{algpseudocode}
\usepackage{graphicx} %
\usepackage[final]{acl}
\usepackage{times}
\usepackage{latexsym}
\usepackage{algorithm}
\usepackage{natbib}
\usepackage{amsmath}
\usepackage{amssymb}
\usepackage{booktabs}

\usepackage{pifont}

\usepackage[T1]{fontenc}

\usepackage[utf8]{inputenc}

\usepackage{microtype}

\usepackage{inconsolata}
\usepackage[textsize=tiny]{todonotes}

\newcommand{\ramya}[1]{\begingroup\color{teal} #1\endgroup}

\usepackage{colortbl}
\usepackage{soul}
\usepackage{xcolor}
\newcommand{\highlight}[1]{{\color{magenta}\{\{#1\}\}}}

\usepackage[skins,breakable]{tcolorbox}
\usepackage{tikz}
\usetikzlibrary{tikzmark,decorations.pathreplacing,calc}

\definecolor{light-purple}{RGB}{151,156,171}
\definecolor{blue-color}{RGB}{40,166,189}
\definecolor{pink-color}{RGB}{237,46,104} 
\definecolor{dark-grey-color}{RGB}{79,91,102}
\definecolor{darkbyzantium}{rgb}{0.36, 0.22, 0.33}
\definecolor{bluebell}{rgb}{0.64, 0.64, 0.82}
\definecolor{airforceblue}{rgb}{0.36, 0.54, 0.66}
\definecolor{response}{RGB}{245,198,165}

\definecolor{greenhl}{RGB}{60, 140, 80}
\definecolor{greenfill}{RGB}{220, 245, 225}
\definecolor{hellowfill}{RGB}{250, 240, 200}
\definecolor{redhl}{RGB}{180, 50, 50}
\definecolor{redfill}{RGB}{250, 220, 220}
\definecolor{bluefill}{RGB}{230, 240, 255}

\definecolor{violethl}{HTML}{DCCFE5}
\definecolor{brickhl}{HTML}{f7cad0}
\definecolor{yellowhl}{HTML}{ffd000}

\definecolor{peachhl}{HTML}{F3DCCA}
\definecolor{bluehl}{HTML}{AACBE5}
\definecolor{limehl}{HTML}{C7E599}
\definecolor{tealhl}{HTML}{C5E4E5}

\newcommand{\promptsubsection}[1]{
\setlength{\parskip}{6pt} \noindent\textbf{{#1}:}
}

\newcommand{\rubricsubsection}[1]{
\setlength{\parskip}{6pt} \noindent\textbf{{#1}:}
}

\newcommand{\examplesubsection}[1]{
\setlength{\parskip}{6pt} \noindent\textbf{{#1}:}
}

\newtcolorbox[list inside=prompt,auto counter,number within=section]{prompt}[1][]{
    colbacktitle=airforceblue,
    colframe=airforceblue,
    fontupper=\footnotesize,
    boxsep=5pt,
    left=0pt,
    right=0pt,
    top=0pt,
    bottom=0pt,
    boxrule=1pt,
    enhanced, 
    breakable,
    skin first=enhanced,
    skin middle=enhanced,
    skin last=enhanced,
    #1,
}

\newtcolorbox[list inside=prompt,auto counter,number within=section]{response}[1][]{
    colbacktitle=response,
    colframe=response,
    fontupper=\footnotesize,
    boxsep=5pt,
    left=0pt,
    right=0pt,
    top=0pt,
    bottom=0pt,
    boxrule=1pt,
    enhanced, 
    breakable,
    skin first=enhanced,
    skin middle=enhanced,
    skin last=enhanced,
    #1,
}

\newtcolorbox[list inside=rubric,auto counter,number within=section]{rubric}[1][]{
    colbacktitle=darkbyzantium,
    colframe=darkbyzantium,
    fontupper=\footnotesize,
    boxsep=5pt,
    left=0pt,
    right=0pt,
    top=0pt,
    bottom=0pt,
    boxrule=1pt,
    enhanced, 
    breakable,
    skin first=enhanced,
    skin middle=enhanced,
    skin last=enhanced,
    #1,
}

\newtcolorbox[list inside=example,auto counter,number within=section]{example}[1][]{
    colbacktitle=dark-grey-color,
    colframe=dark-grey-color,
    fontupper=\footnotesize,
    boxsep=5pt,
    left=0pt,
    right=0pt,
    top=0pt,
    bottom=0pt,
    boxrule=1pt,
    enhanced, 
    breakable,
    skin first=enhanced,
    skin middle=enhanced,
    skin last=enhanced,
    #1,
}

\usepackage{hyperref}
\usepackage{fontawesome5}

\usepackage{xspace}

\DeclareRobustCommand{\cnum}[1]{%
  \tikz[baseline=(c.base)]{%
    \node[circle,fill=black,inner sep=1pt,
          font=\scriptsize\bfseries\color{white}] (c) {#1};%
  }%
}

\newcommand{\Agent}{%
  \setlength{\fboxsep}{2pt}%
  \colorbox{bluehl}{{\textsc{agent}}}%
  \xspace
}

\newcommand{\Plot}{%
  \setlength{\fboxsep}{2pt}%
  \colorbox{brickhl}{{\textsc{plot}}}%
  \xspace
}

\newcommand{\Perspective}{%
  \setlength{\fboxsep}{2pt}%
  \colorbox{limehl}{{\textsc{perspective}}}%
  \xspace
}

\newcommand{\Persp}{%
  \setlength{\fboxsep}{2pt}%
  \colorbox{limehl}{{\textsc{persp}}}%
  \xspace
}

\newcommand{\Setting}{%
  \setlength{\fboxsep}{2pt}%
  \colorbox{violethl}{{\textsc{setting}}}%
  \xspace
}

\newcommand{\SocialAtmosphere}{%
  \setlength{\fboxsep}{2pt}%
  \colorbox{peachhl}{{\textsc{social atmosphere}}}%
  \xspace
}

\newcommand{\SocAtm}{%
  \setlength{\fboxsep}{2pt}%
  \colorbox{peachhl}{{\textsc{soc. atm.}}}%
  \xspace
}

\newcommand{\Style}{%
  \setlength{\fboxsep}{2pt}%
  \colorbox{tealhl}{{\textsc{style}}}%
  \xspace
}

\newcommand{\genie}[0]{\textsc{Genie}\xspace}

\title{\raisebox{-2ex}{\includegraphics[width=1cm]{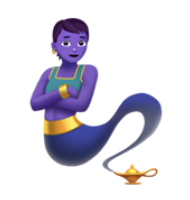}}\genie: A Fine-Grained Measure for Novelty}

\author{Ramya Namuduri$^1$ \ \ \ \ 
Manya Wadhwa$^2$ \ \ \ \ 
Anshun Asher Zheng$^1$\\
\textbf{Greg Durrett}$^2$ \ \ \ \ 
\textbf{Junyi Jessy Li}$^1$ \\ 
$^1$The University of Texas at Austin \quad $^2$New York University
\\
\texttt{ramya.namuduri@utexas.edu}
\AND
\href{https://github.com/AlliteraryAlligator/GENIE/tree/main}{\faGithub\ Code} \quad 
\href{https://huggingface.co/collections/AlliteraryAlligator/genie}{\raisebox{-1ex}{\includegraphics[height=1.5em]{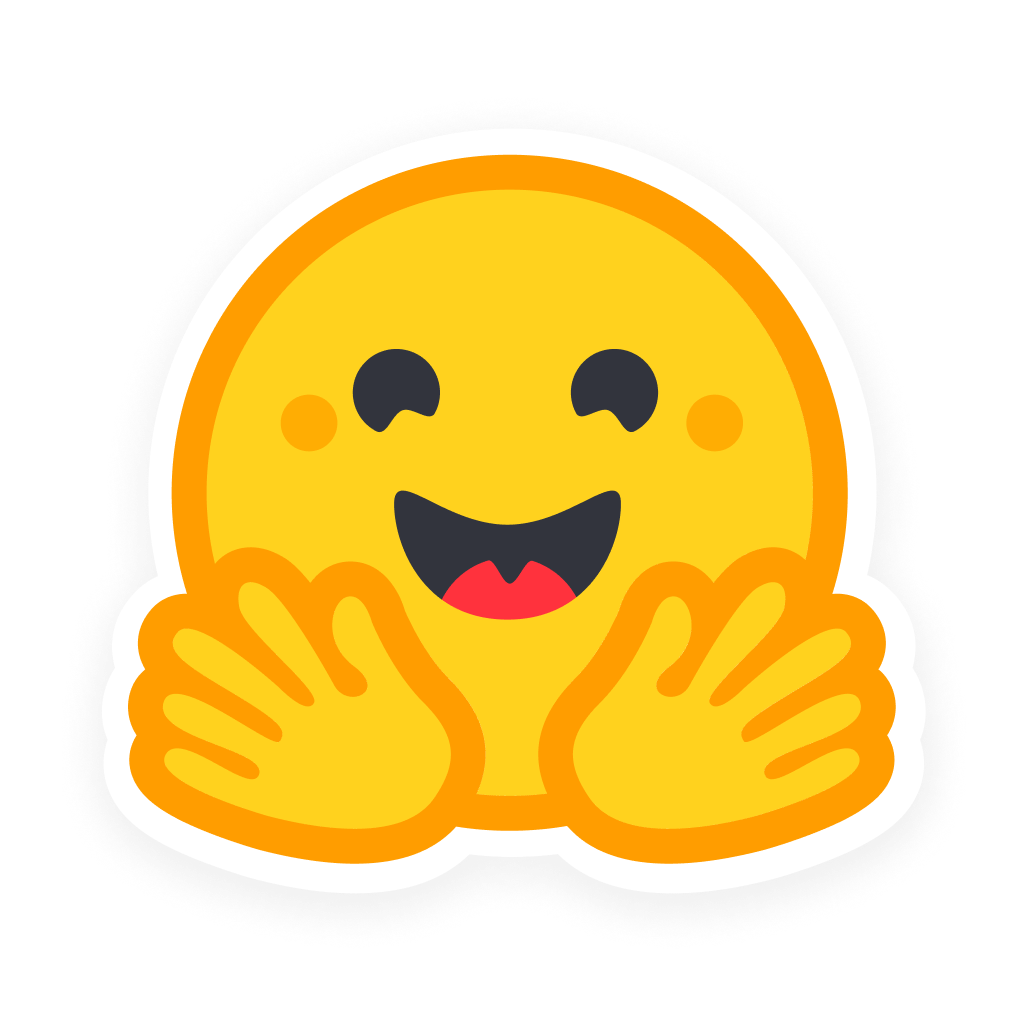}} Data} \quad
\href{https://alliteraryalligator.github.io/GENIE/}{\faLink\ Project Page}
}

\begin{document}

\maketitle

\begin{abstract}

Large Language Models have consistently demonstrated a lack of creativity and diversity across tasks.
Prior work has focused on addressing \textit{whether} models are capable of generating creative outputs. Here, we aim to consider \textit{novelty} and investigate what makes model-generated content novel or not novel in a task-specific manner. We propose a fine-grained evaluation metric \genie to measure the novelty of responses along task-specific \textit{features} with respect to a population of responses.
We show that unlike \genie, holistic metrics struggle to capture the high-dimensionality of novelty and do not provide insight on which properties they target. 
Finally, we use \genie to measure the effectiveness of mitigation methods that address creativity to better understand where these methods can improve novelty.

\end{abstract}

\section{Introduction}

Can LLMs produce creative work? Prior work has discovered that models struggle to generate diverse and creative outputs given various tasks \cite{jiang2026artificial}, using lexical-semantic approaches \cite{zhang2025noveltybench, lu2025ai, ismayilzada2025creative}, creativity tests \cite{Wenger2025WereDW, wadhwa2026create}
and human experts \cite{Chakrabarty2023ArtOA}. 
Model creativity has been tested on a variety of tasks using a multitude of metrics.
However, past work has not isolated precisely \textit{where models lack creativity} on specific tasks.

\begin{figure}
    \centering
   \includegraphics[scale=0.38,trim=0 21cm 25cm 0]{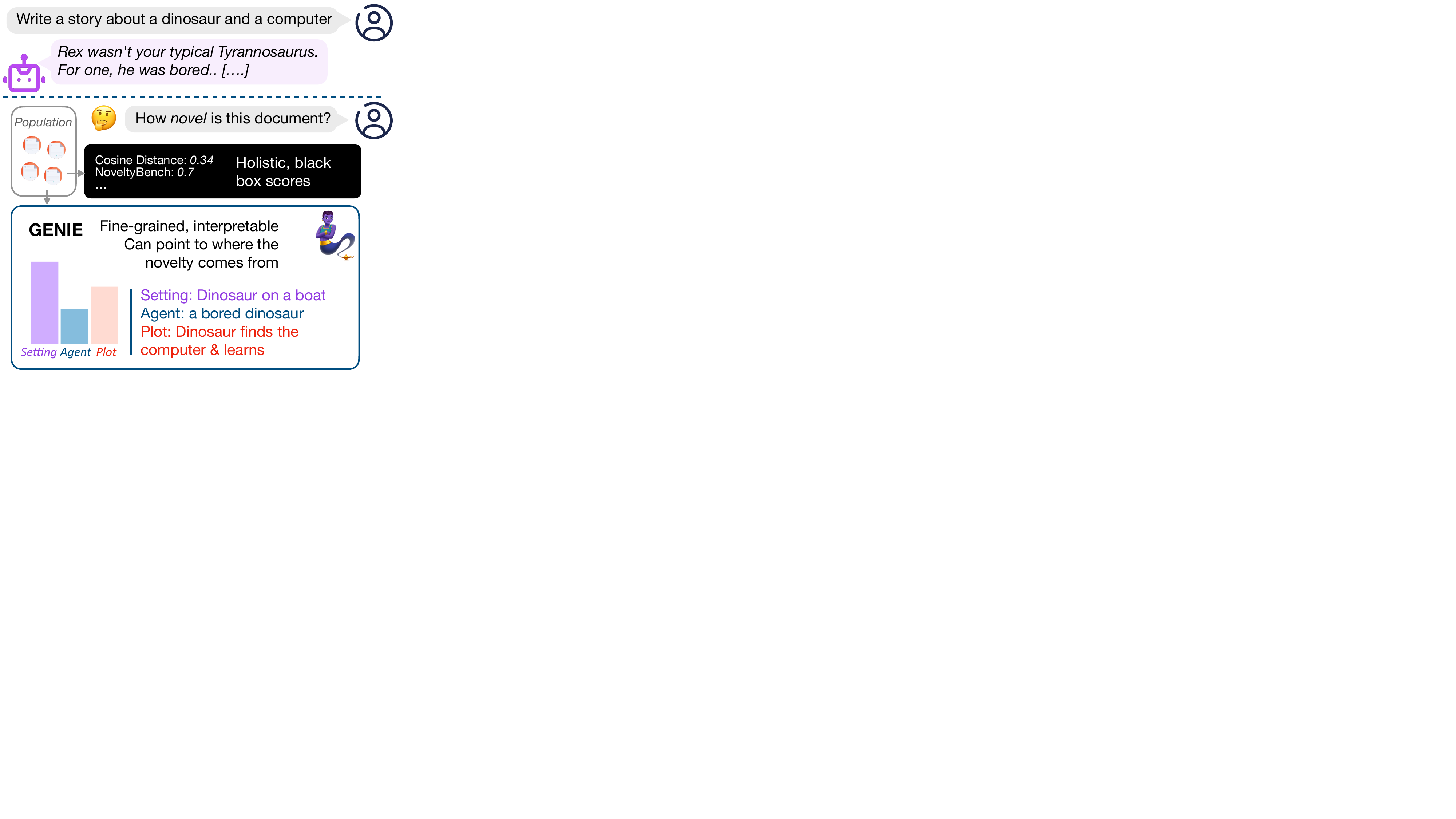}
    \caption{High level motivation for \genie. Existing holistic metrics do not capture the high-dimensionality of novelty while \genie's fine-grained approach provides explainability.} %
    \label{fig:high_level}
\end{figure}

Novelty is a key component of creativity \cite{boden1991creativemind}, where the uniqueness of a response can be attributed to different reasons specific to the task. 
Features such as \textit{plot} and \textit{character development} are relevant to narrative writing, while \textit{motivation} and \textit{methodology} are applicable to scientific discovery.
Two responses can be strikingly original, and yet differ vastly in what makes them novel. Similarly, a response might appear novel in some ways and commonplace in others.
For instance, Figure \ref{fig:high_level} illustrates how an LLM-generated response can concurrently have unique and mundane features.
However, this fine-grained novelty is not captured by many existing holistic creativity metrics. Cosine distance against other responses, and other creativity metrics \citep{zhang2025noveltybench, chakrabarty2025aislop, fein2025litbenchbenchmarkdatasetreliable, Shaib2024DetectionAM}, although informative, do not describe why an instance is more or less novel.

We propose \genie (\textbf{G}ranular \textbf{E}valuation of \textbf{N}ovel \textbf{I}deas with \textbf{E}xplainability), 
a framework that quantifies the fine-grained nuances that differentiate equally novel responses by extracting \textit{features} present in the responses, with respect to a pre-defined population.
Inspired by alternative semantics of questions literature  \citep{hamblin1957language, karttunen1977syntax, lahiri2002questions},
we do so by generating and mapping questions to automatically discovered features relevant to the task,  
and measuring fine-grained novelty through the similarities of their answers. 

\genie's focus on automatically extracted \textit{features} isolates novelty from other components of creativity, such as value (i.e., quality; \citet{boden1991creativemind}).
This is in contrast to some existing metrics that
subsume the quality of responses \cite{chakrabarty2025aislop} or the predicted aesthetic value \cite{fein2025litbenchbenchmarkdatasetreliable}. By tethering the feature extraction to the \textit{task}, \genie's approach 
is inherently domain agnostic. 

\genie's level of abstraction (using questions and answers)
allows it to treat models of varying instruction-following capabilities comparably, specifically with regards to models with poor instruction-following capabilities that exhibit greater diversity despite lower quality \cite{West2025BaseMB, le-bronnec-etal-2024-exploring, Hamilton2024DetectingMC}.

Furthermore, determining the uniqueness of responses holistically without a reference point can be subjective. Prior work has used the pretraining corpus \cite{lu2025ai}, other model-generated \cite{jiang2026artificial}, human \cite{Wenger2025WereDW} or self-generated \citep{Zhang2025VerbalizedSH, zhang2025noveltybench} responses as reference points for their measurements. In this work, we make the references explicit: \genie measures novelty \textit{with respect to a population}; this population is defined by the user.

We explore the creative writing domain. \genie independently found $7$ fine-grained features pertaining to this task, which roughly correspond to theoretically defined taxonomies \citep{Hamilton2025NarraBenchAC}. We constructed and used a population of responses generated by 21 LLMs to evaluate the \textit{sensitivity} of \genie on pairs of LLM-generated responses that minimally differ along these features. 
\genie successfully detects differences in fine-grained novelty across such pairs. \genie also demonstrates \textit{robustness} on pairs that only differ via paraphrasing; it does \emph{not} detect a significant difference along any feature in these cases. Additionally, we performed human studies to intrinsically evaluate various components that comprise \genie. %

\genie's fine-grained nature provides a diagnostic lens into existing holistic metrics 
used to measure novelty, diversity and creativity.
We found that they either capture extraneous properties that sometimes spuriously correlate with novelty, or are lexically sensitive. Furthermore, many existing mitigation methods, such as alternate prompting \cite{Zhang2025VerbalizedSH, huot2025agents} and decoding strategies \cite{peeperkorn2025mind}
for improving creativity have been evaluated on these holistic metrics. We demonstrate how \genie provides a more explainable evaluation, shedding light into \emph{which} features, if any, these methods actually improve.

\begin{figure*}[t!]
    \centering
    \includegraphics[scale=0.25,trim=0cm 17cm 6cm 0cm]{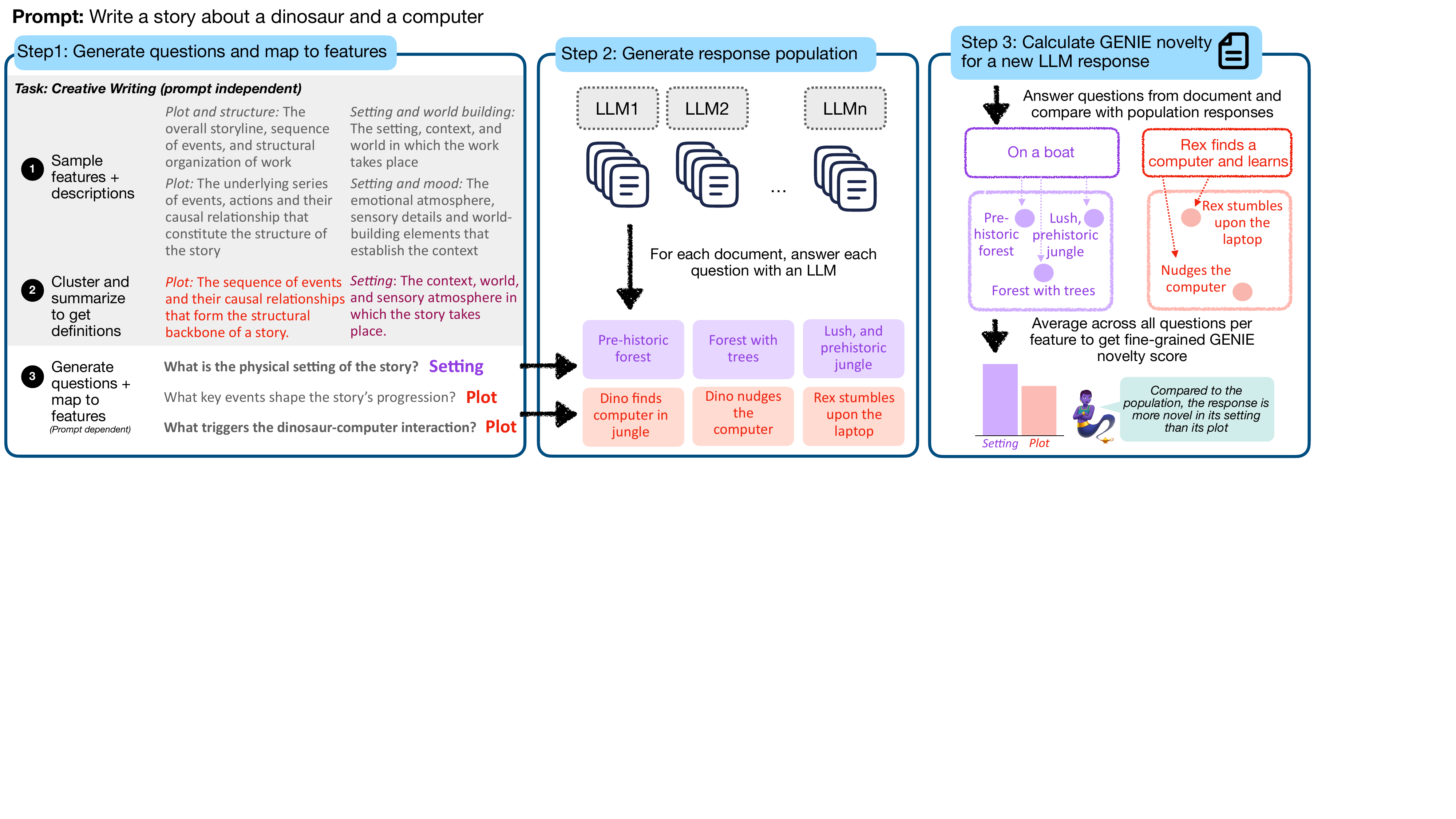}
    \caption{\genie captures fine-grained novelty of a target response with respect to a population of responses using automatically discovered task-specific features. Here, \genie discovers features and their descriptions for creative writing. We generate questions for a given prompt $p$ (Step \cnum{3}) and map each question to a feature.
   Step 2 computes answers to questions from the population. Finally, \genie measures the novelty of a target response with respect to this population, along each feature (e.g. setting or plot) using the average pairwise answer dissimilarity. A complete example is provided in Appendix \ref{appendix:genie-example}. %
   }
    \label{fig:genie-process}
\end{figure*}

\section{Related Work}

A growing line of work examines whether LLMs are creative, drawing on theoretical frameworks from cognitive science and computational creativity. Many studies operationalize Boden's theories of creativity \cite{franceschelli2024creativity, ismayilzada2025creative, schapiro2025combinatorial, nagarajan2025roll, wadhwa2026create} while others adapt psychometric tests such as Torrance Tests of Creative Thinking \cite{Chakrabarty2023ArtOA}. Past work has also focused on the homogenizing effects of LLMs when looking at a population of LLM outputs \cite{Wenger2025WereDW, jiang2026artificial} and attributed this to post-training alignment methods \cite{padmakumar2024does, West2025BaseMB, Lanchantin2025DiversePO}. There are also domain specific benchmarks that evaluate creativity in writing \cite{fein2025litbenchbenchmarkdatasetreliable, chakrabarty2025aislop, xu2025echoes, fein2025litbenchbenchmarkdatasetreliable, zhang2025noveltybench}. Most of this work asks whether models are creative; instead we ask where a given response is novel and decompose this judgment along task-specific features.

Recent work has proposed explicit metrics and benchmarks for evaluating novelty and creativity in text generation. One common approach is to use the pretraining corpus as reference to quantify originality as a fraction of n-grams \cite{lu2025ai, padmakumar2026measuring, merrill-etal-2024-evaluating}. NoveltyBench \cite{zhang2025noveltybench} operationalizes diversity by checking for if an output belongs to one of the preexisting clusters of unique responses. Past work also trains reward models on human preference data to score creative writing directly \cite{chakrabarty2025aislop, fein2025litbenchbenchmarkdatasetreliable, cao2026evaluating}. Across all the prior work, novelty is reduced to a scalar where the reference is implicit or calculated with surface units (e.g., n-grams or distributional semantics).
\textsc{QUDsim} \cite{namuduri2025qudsim} decomposes similarity along discourse-level questions under discussion (QUDs) \citep{roberts2012information,VellemanBeaver2016QUD, beaver2017questions}. 
This established the effectiveness of measuring the similarity of answers to the same questions across documents. However, QUDs capture discourse,
whereas \genie decomposes novelty into 
task-level features
against an explicit population, separating content from quality and aesthetics of the output being evaluated.

\section{The \genie Measure of Novelty}

Consider a population of model-generated documents $\mathcal{D}_{pop} = \{d_1, ..., d_m\}$ generated from prompt $P$. The novelty of a target document $d_t \notin \mathcal{D}_{pop}$ is defined as its relative dissimilarity $d$ for all $d\in \mathcal{D}_{pop}$ along various \textit{features}. 
Intuitively, unique responses share fewer commonalities across features with other responses. 
\genie is explainable, because each feature is associated with natural language definitions; when unpacked, further explainability can be obtained by pinpointing prompt-specific questions under that feature, as well as their natural language answers whose similarity we calculate.

\subsection{Features}
Features are task-specific axes that aim to maximize coverage over different characteristics of the task. For instance, a task like creative writing might include features such as \textit{plot}, \textit{setting}, and \textit{perspective}. On the other hand, \textit{motivation} and \textit{methodology} might be relevant to evaluating the novelty of scientific discovery.
Manually defining these features for tasks can be tedious, subjective, or infeasible (due to lack of expertise).
Instead, we automatically \textit{discover} features. 

We prompt an LLM to form features with descriptions given a task (e.g. creative writing) $k$ times. The resulting set is clustered to find commonly recurring features. Concretely, we discover $N$ clusters and label them as $\mathcal{F}=\{f_1, ..., f_N\}$ to form the set of \textit{features}. Step \cnum{1} and \cnum{2} in Figure \ref{fig:genie-process} demonstrate how we discover features for the creative writing task.  
Each feature $f$ is a function $f: \mathcal{D}\to \mathcal{V}_{f}$ that maps a document to its feature value, where $\mathcal{D}$ the set of all documents and $\mathcal{V}_{f}$ is the set of all possible values the feature can take.

\subsection{\genie Scoring}
\label{sec:genie-scoring}
\genie elicits values for each feature $f\in\mathcal{F}$ in a document (i.e., $f(d)$) via question generation and answering. Here, question generation is an intermediate process that grounds $\mathcal{F}$ to prompt-specific dimensions, allowing feature values to be more reliably extracted by answering the questions. This paradigm takes the alternative semantics view of questions \citep{hamblin1957language, karttunen1977syntax, lahiri2002questions}, where hierarchically organized questions are used to partition the possible worlds into subspaces that ``answer the same questions'', in our case, ``values of the same feature $f$'', hence allowing us to measure novelty \emph{for each feature}.

\vspace{-0.3em}
\paragraph{Prompt-Specific Decomposition}
\label{sec:question-generation} 
Given a prompt $p \in P$, we generate a set of questions $\mathcal{Q}=\{q_1, ..., q_j\}$ (see Step \cnum{3} in Figure \ref{fig:genie-process}). The set $\mathcal{Q}$ is formed without knowledge of $\mathcal{F}$ so that the resulting questions are specific to $p$, while not overfitting to the taxonomy of discovered features.

\vspace{-0.3em}
\paragraph{Feature Mapping}
\label{sec:feature-map} 
We then map each question in $\mathcal{Q}$ to exactly one \textit{feature} $f\in \mathcal{F}$ (Step \cnum{3} in Figure \ref{fig:genie-process}). The subset of questions that map to feature $f$ is denoted by $\mathcal{Q}_f \subseteq \mathcal{Q}$. We can represent the feature value $f(d) \in \mathcal{V}_f$ as the set of answers to $\mathcal{Q}_f$ for document $d$ (see Step 2 in Figure \ref{fig:genie-process}).

\vspace{-0.3em}
\paragraph{\genie}
The \genie novelty of $d_t$ is defined as a \textbf{vector} that consists of the individual novelty scores along each feature $f \in \mathcal{F}$:
\[
\mathcal{\mathbf{G}}(d_t) = [\text{G}_{\text{feat}}(d_t, f)]_{f \in \mathcal{F}}
\]

The novelty of the target document along $f$ is defined as the average novelty across all questions $q \in \mathcal{Q}_f$:

\[
\text{G}_{\text{feat}}(d_t, f) = \frac{1}{|\mathcal{Q}_{f}|} \sum_{q_i \in \mathcal{Q}_{f}} \text{G}_{\text{quest}}(d_t, q_i)
\]

\noindent Where $\text{G}_{\text{quest}}(d_t, q_i)$ (i.e., the novelty of $d_t$ along $q$)
is defined as the average pairwise dissimilarity ($1-$ similarity) between the target answer $a$ and the corresponding answers in each population document (see Step 3 in Figure \ref{fig:genie-process}). Let $a_{d_t}$ and $a_{d_p}$ be answers to question $q$ found in target document $d_t$ and population document $d_p \in \mathcal{D}_{pop}$, respectively.
\vspace{-0.5em}
\[
\text{G}_{\text{quest}}(d_t, q) = \frac{1}{|\mathcal{D}_{pop}|} \sum_{d_p \in \mathcal{D}_{pop}} \mathrm{dissim}(a_{d_t}, a_{d_p})
\]
\vspace{-1em}

Intuitively, $d_t$ will get a higher \genie score on feature $f$ if that feature is distinctive compared to how $f$ is realized among the population documents. In the example shown in Figure \ref{fig:genie-process}, $d_t$ receives a higher score on \textit{setting} than \textit{plot} due to its higher dissimilarity with the population of \textit{settings}. We keep \genie as a vector rather than an aggregate score to preserve fine-grained explainability. Additionally, the novelty of a response can be independently attributed to multiple features.

\section{\genie for Creative Writing} \label{sec:genie_instantiation}
In this paper, we instantiate \genie on the creative writing task, in particular long-form story generation. We choose this task for several reasons: (1) it is known to be multi-faceted with fine-grained features 
such as \textit{plot}, \textit{setting}, etc.\ that are researched in prior work \cite{Hamilton2025NarraBenchAC, Russell2026StoryScopeII}.
These existing features can validate the features that we automatically derived.  (2) Even though LLMs can generate coherent text, long-form stories generated by LLMs are known to be uncreative or homogeneous.
Thus, it is a particularly valuable test bed for fine-grained, explainable insights into \textit{where} the (lack of) novelty appears, especially as existing metrics do not currently do so.

\subsection{Tasks} \label{sec:tasks}
We curate a total of \textasciitilde 10,000 documents using 39 models, across 50
open-ended writing prompts ($P$) gathered from two sources: (1) 25 synthetic prompts generated by \texttt{gpt-4.1-mini} following the format \textit{Write a story about X and Y} where \textit{X} and \textit{Y} are spatially or temporally distant concepts (Prompt \ref{prompt:synthetic-concepts}) 
and (2) 25 prompts sampled from the LitBench \cite{fein2025litbenchbenchmarkdatasetreliable} subset of the \texttt{r/WritingPrompts} dataset \cite{fan2018hierarchical}. Examples of the prompts in our dataset are in Appendix \ref{appendix:prompt-examples}.
A power analysis reveals the number of prompts is a sufficiently large sample size for our experiments to achieve statistical significance (Appendix \ref{appendix:population-sampling-process}).

\subsection{Features} \label{sec:feature_discovery}
We gathered features describing the creative writing task from prompting \texttt{gpt-4.1} $10$ times 
(Prompt \ref{prompt:feature-discovery}) eliciting a definition for each,
and clustered the responses using embeddings computed with \texttt{text-embedding-3-small}.\footnote{HBDSCAN, Agglomerative and K-means clustering resulted in similar clusters; we set $k$ to the average number of features in each sample.} 
The features discovered include: \textsc{setting, plot, characterization, mood/tone, point of view, language, theme}, which closely correspond to the taxonomy described in \citet{Hamilton2025NarraBenchAC}. We therefore adopt their terminology and map our discovered features accordingly. In this work, we specifically focus on \Agent, \Perspective, \Plot, \Setting, \SocialAtmosphere,\footnote{We renamed \textit{Social Network} as used in NarraBench to \textit{Social Atmosphere} due to the former's loaded meaning.} and \Style to form the set $\mathcal{F}$. Details on how the features and definitions we discovered were mapped are included in Appendix \ref{appendix:feature-discovery}.

\subsection{Question Generation}
For each prompt $p \in P$, we (a) generate the set of questions $\mathcal{Q}$ (\S\ref{sec:question-generation}), (b) perform filtering, and (c) map each question to the most representative feature in $\mathcal{F}$. Each stage of the pipeline uses \texttt{gpt-4.1-mini}. The prompts are in Appendix \ref{appendix:question-prompts}.

\vspace{-0.3em}
\paragraph{Filtering} Questions undergo a one-step filtering stage
to exclude questions that are polar, hypothetical, conditional, subjective or those that address multiple features.
Since the goal of using questions is to extract features and view them as independently as possible (i.e. the novelty of \Setting is not dependent or influenced by the novelty of \Style).

\vspace{-0.3em}
\paragraph{Feature Mapping}
We then map $\mathcal{Q}$ to $\mathcal{F}$ as described in \S \ref{sec:feature-map}. This results in $\mathcal{Q}_f$ for all six creative writing features. 
Table \ref{question-stats} shows statistics on the distribution of questions generated. \Agent and \Plot are the largest features in question coverage, owing to the richness of the content that can be considered as either the storyline or character development. \Plot requires more questions to cover the conflict, turning points, obstacles, plotline, etc. On the other hand, \Perspective questions are largely limited to extracting \textsc{point of view} related information.

\begin{table}[ t!]
\small
\renewcommand{\arraystretch}{1}
\renewcommand{\tabcolsep}{1.0mm}
\centering
\begin{tabular}{lcc}
\toprule
\multicolumn{1}{c}{Feature} &\multicolumn{1}{c}{\bf Count} &\multicolumn{1}{c}{\bf Avg. per Prompt} \\

\midrule
\Agent & 254 & 4.23 \\ 
\Perspective & 79 & 1.61\\ 
\Plot  & 369 & 6.15\\ 
\Setting  & 140 & 2.86 \\ 
\SocAtm & 82 & 1.82\\ 
\Style & 108 & 2.00 \\ 
\midrule
Total & 1032 & 17.20 \\
\bottomrule
\end{tabular}
\vspace{-0.5em}
\caption{Distribution of questions generated per feature and per prompt. \Plot and \Agent features are represented by more questions, per prompt, due to their higher complexity relative to \Perspective. } \label{question-stats}
\vspace{-1em}
\end{table}

\subsection{Constructing the Population and Target Set} 
\paragraph{Models} 
Using prompts from Section \ref{sec:tasks}, we build a \textit{population} consisting of 21 models ranging across various families, sizes and stages of training (Table \ref{population-models}).
We consider 18 \textit{target} models (Table \ref{target-models}).
The target models roughly correspond to newer iterations of the population models so that \genie
can reflect novelty differences as models evolve.

\vspace{-0.3em}
\paragraph{Responses} For each prompt, we sample 5 responses from every model, across the population and target models (see Step 2 in Figure \ref{fig:genie-process}). All responses are post-processed to truncate incoherent or irrelevant responses, and reasoning. We then use ArmoRM \cite{ArmoRM} to compute coherence and further discard documents that do not meet a minimum threshold of 2.0.\footnote{Completely incoherent documents can have artificially high novelty and thus must be filtered out. This threshold is chosen after manual inspection on a sample of documents.}

The \textbf{target} set for which we calculate novelty consists of 4500 documents. We use a pilot dataset and conduct a power analysis to determine the statistically significant population size required to reach 98\% confidence with a $0.1$ margin of error. The resulting \textbf{population} contains 3404 documents.
Details about the sampling process are in Appendix \ref{appendix:population-sampling-process}.

\subsection{Dissimilarity Function}
\label{sec:distance-function}
To calculate $\mathrm{dissim}(a_{d_t}, a_{d_p})$, we judge similarity using \texttt{Qwen2.5-32B-Instruct} on a 4-point Likert scale using the rubric shown in \ref{rubric:sim}
and compute $1-\mathrm{similarity}$. Two answers are considered highly similar if they are interchangeable, despite slightly different wording. On the other hand, dissimilar answers have very low overlap in the details described, and sometimes can contradict each other. Pairs that are difficult to judge often have overlapping content or scope.
Some questions are unanswerable by certain documents (``unspecified'', ``not applicable'', etc.). 
We exclude these answers to prevent %
\genie scores from rewarding unanswerability.

\section{Evaluating \genie} \label{sec:eval_genie}

We intrinsically evaluate components of \genie, such as question generation, feature mapping and the dissimilarity function, with the help of human annotators (recruitment details are provided in \ref{appendix:recruitment}). We then conduct a study to measure \genie's \textit{sensitivity} to specific features and its \textit{robustness} to surface-level variation.

\subsection{Intrinsic Validation}

\paragraph{Question Generation} \label{sec:qg-eval-results} We run a small scale human study with two annotators
to evaluate the reliability of using an LLM for generating questions given a prompt (\S\ref{sec:question-generation}). The annotators are give $50$ questions and are asked to evaluate the validity of each question using the criteria given in Appendix \ref{rubric:q-validity}. The inter-annotator agreement (Cohen's kappa) is $\mathbf{0.59}$  (moderately high, \citet{arsteinInterCoderAgrCL}).
Some common source of disagreements are listed in Appendix \ref{appendix:validity-mapping-disagreement}.
The question generator \texttt{gpt-4.1-mini} achieves an F1 score of $\mathbf{0.71}$ and $\mathbf{0.74}$ as evaluated by two annotators.

\vspace{-0.3em}
\paragraph{Feature Mapping} \label{sec:mapping-eval-results} Since reliable question-to-feature mapping is essential for feature coverage and low noise, we validate the mapping step in \S\ref{sec:feature-map}.
Two annotators classified a set of 50 questions into one of the six features, or none, as given in \S\ref{sec:genie_instantiation} using the criteria in Appendix~\ref{rubric:q-map}. Cohen's kappa
was high at $\mathbf{0.75}$, and as shown in Table~\ref{table:q-f1}, \texttt{gpt-4.1-mini} achieves high F1 scores across all features against both annotators.

\vspace{-0.3em}
\paragraph{Similarity} To validate the LLM dissimilarity function, we asked human annotators to judge the similarity of $3,000$ randomly sampled pairs of answers according to guidelines in Appendix \ref{rubric:sim}. The annotation set is selected to maximize coverage across features and target-population model pairs. Each pair of answers is annotated by two annotators. Table \ref{table:krippendorff} reports per-feature
agreement between annotators, as well as between annotators and \texttt{Qwen2.5-32B-Instruct} (the LLM used for distance judgments). Measured with Krippendorff-$\alpha$, inter-annotator agreement is $\mathbf{0.59}$, and higher $\mathbf{0.66}$ for annotators vs.\ LLMs.

\begin{table}[t!]
\small
\renewcommand{\arraystretch}{1}
\renewcommand{\tabcolsep}{1.0mm}
\centering
\begin{tabular}{lcc}
\toprule
\multicolumn{3}{c}{\bf Feature Mapping Model Accuracy}\\
\midrule
\multicolumn{1}{l}{Task} & \multicolumn{1}{c}{\bf LLM v. Ann1} & \multicolumn{1}{c}{\bf LLM v. Ann2} \\
\midrule
\Agent & 0.67 & 1.00\\
\Perspective & 0.73 & 0.92\\
\Plot & 0.81 &0.90\\
\Setting & 0.62 &0.67\\
\SocAtm & 0.83 &1.00\\
\Style & 0.77 &0.83\\
\bottomrule
\end{tabular}
\vspace{-0.5em}
\caption{Model (\texttt{gpt-4.1-mini}) accuracy for question-to-feature mapping.
}
\label{table:q-f1}
\vspace{-1em}
\end{table}

\subsection{\genie Sensitivity to Feature Edits}
\label{sec:genie-causal-intervention}
We
conduct a causal intervention to investigate the \textit{sensitivity} of \genie to differences between responses along specific features. \textbf{Sensitivity} is defined as the ability to detect a statistically significant change in \genie's observed novelty for feature $f$ when an intervention is performed along $f$. \textbf{Robustness} refers to statistical equivalence in the observed novelty for feature $f$ when a lexical edit, such as paraphrasing, is performed.
An ideal metric for measuring content-level novelty detects all interventions (high sensitivity), while not responding to paraphrases (high robustness).

To test for sensitivity, we surgically edit documents along individual features using population-agnostic synthetic alternatives, while maximally preserving other features.
To test for robustness, we paraphrase documents in a semantics-preserving manner, altering surface form while preserving all content features.

\subsubsection{Experimental Setup}

\paragraph{Feature Intervention}
We first define an edit function $E: \mathcal{D} \times \mathcal{F} \times \mathcal{V} \rightarrow \mathcal{D}$ that produces a minimally edited document $d' = E(d, f_i, \tilde{v}_{f_i})$, where $f_i \in \mathcal{F}$ is the feature being intervened upon
and $\tilde{v}_{f_i}$ is the new proposed value of $f_i$.
The original document $d$ and the edited document $d'$ differ along the value of $f_i$
while maximally preserving all else.
Concretely, we synthetically generate an alternate value $\tilde{v}_{f_i}$ for the target feature using an LLM, conditioned on $d$ and the prompt $p$ but independent of any population (Prompt~\ref{prompt:alt-features}); $d'$ is then the rewrite of $d$ that reflects $\tilde{v}_{f_i}$, %
with all \textit{other} feature values held fixed (i.e., $\forall f \neq f_i: f(d) = f(d')  $). 
This helps ensure that there is a meaningful difference along $f_i$ between $d$ and $d'$ without a predefined notion of relative novelty.
We sample alternate features using \texttt{gpt-5.4-mini} (Prompt \ref{prompt:alt-features}) and implement the edits using \texttt{gpt-4.1-mini} (Prompt \ref{prompt:editing-features}).\footnote{Stronger models are not suitable for generating minimal edits required for this study \cite{Laban2026LLMsCY}.} Examples of alternate features are provided in Appendix \ref{example:alt-features}.

\vspace{-0.3em}
\paragraph{Dataset}
We randomly sample $100$ instruct-model generated documents from the target dataset. 
For each document and along each feature, we generate 3 alternatives and perform edits to produce 3 edited documents. This results in $1800$ total pairs of original and edited documents, of which we filtered out $206$ due to malformed responses (details provided in Appendix \ref{appendix:intervention-style}).

\vspace{-0.3em}
\paragraph{Paraphrasing Documents}
We form $3$ paraphrases per document, resulting in $300$ additional pairs.
Semantically equivalent paraphrasing should not alter feature values.
Thus, a metric that finds a significant difference between paraphrased texts may be capturing other properties (lexical, aesthetic qualities, etc.). 

\vspace{-0.3em}
\paragraph{Intervention Validation}
We validate the edit function along three main criteria: (1) The proposed alternative feature $\tilde{v}_{f_i}$ is \textbf{valid} if $\tilde{v}_{f_i} \in \mathcal{V}_{f_i}$ (i.e. the proposed alternative must reflect the intended intervention);  (2) An edit is \textbf{effective} if $f_i(d')=\tilde{v}_{f_i}$ (i.e. the edits fully implement the proposed alternative such that it is extracted as the new feature value;
(3) Edits are \textbf{minimally invasive} if $\forall f \neq f_i: f(d') = f(d)$.  For instance, editing a story to be written from a different \Perspective can lead to portions of the narrative being omitted for cohesion, thus dramatically impacting which characters or plot points are included. 

The authors annotated a set of 270 pairs of documents along this criteria on a 4-point Likert scale using the rubric described in Appendix \ref{rubric:intervention}). Table \ref{table:randolph-intervention} shows the inter-annotator agreement, measured using Randolph's-$\kappa$ \citep{randolph}.\footnote{This task did not have a fixed rating distribution prior to annotations.}
Agreement is high for judging validity, and moderately high for determining effectiveness. 
Judging invasiveness is subjective.
Thus, we measure the proportion of alternatives that are valid, edits that are effective and minimally invasive by computing the majority vote across annotators (Table \ref{table:intervention-accuracy}). We discover that the edit function generates reliably valid and effective interventions. We note that these edits are invasive for \Plot, \SocialAtmosphere and \Perspective interventions, suggesting inherent inter-feature dependence.

\begin{figure}
    \centering
    \includegraphics[width=1\linewidth]{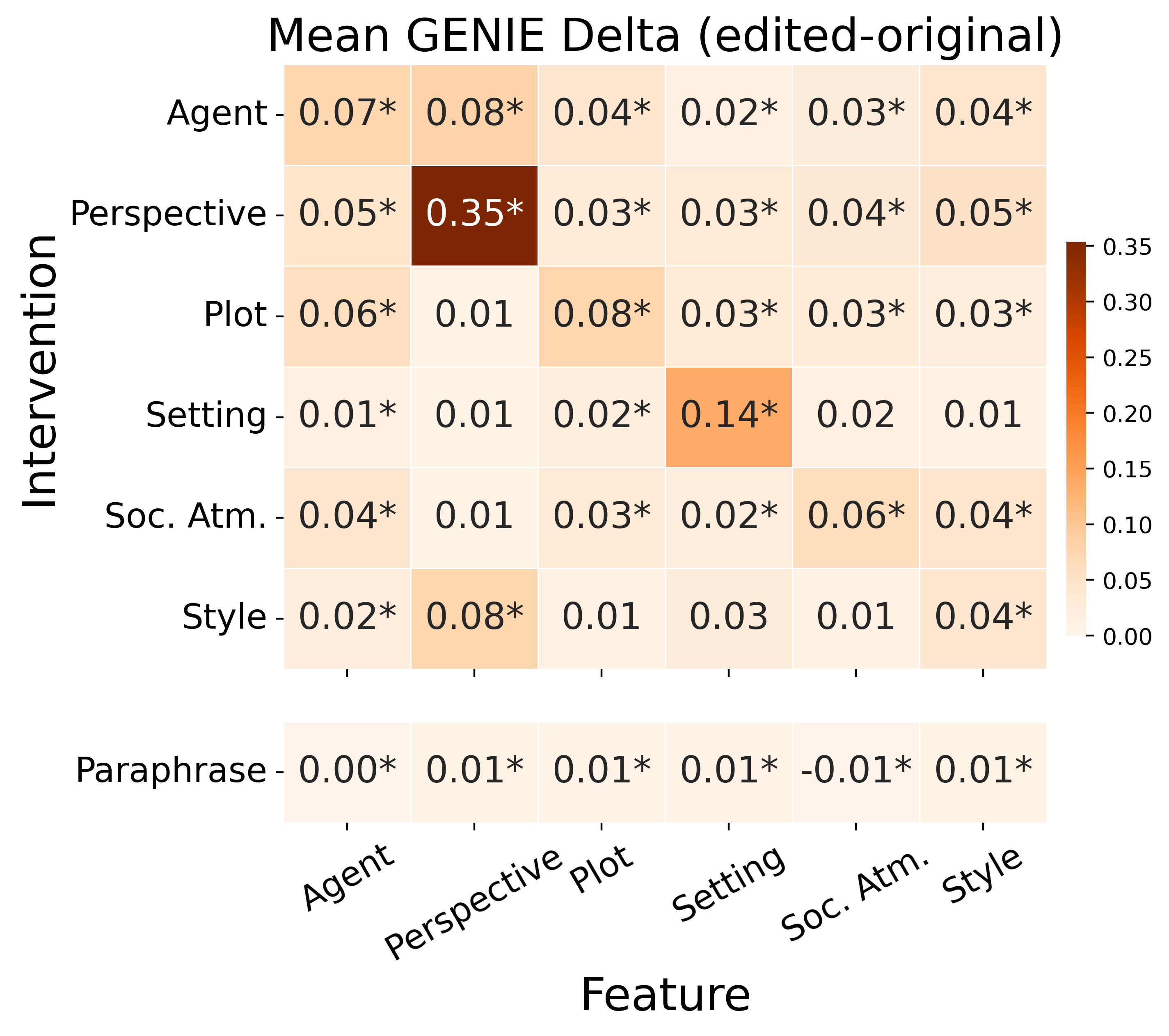}
    \vspace{-2em}
    \caption{Mean deltas registered by \genie for each intervention along each feature. (*): Statistical significant if $p<0.008$ (Bonferroni corrected).
    \genie is most sensitive to \Perspective and \Setting interventions, and registers the largest delta for feature $f_i$ when $f_i$ is intervened upon, except for \Agent and \Style.
    }
    \vspace{-1em}
    \label{fig:genie-all-scores}
\end{figure}

\subsubsection{Sensitivity}
We test whether intervening on $f_i$ produces a systematic change in \genie's score along $f_i$, for all $f_i\in\mathcal{F}$. We perform a two-sided paired bootstrap test to find significance between pre- and post-intervention score distributions. For paraphrases, equivalence is tested using two one-sided tests. %
Refer to Appendix \ref{appendix:stats-test}.

\vspace{-0.3em}
\paragraph{\genie is sensitive to interventions}
Figure \ref{fig:genie-all-scores} shows the mean deltas registered by \genie pre- and post- intervention, per feature.
\genie accurately registers a statistically significant difference for all feature interventions, which shows that it can detect differences in novelty. Moreover, the heatmap indicates that when we intervene on \Perspective, \Plot, \Setting and \SocialAtmosphere, \genie not only detects the change, but also registers the largest mean delta along the intervened feature relative to all other feature scores.

\vspace{-0.3em}
\paragraph{\genie is robust against paraphrases} 
The final row in Figure \ref{fig:genie-all-scores} shows the mean deltas for paraphrased pairs. The mean deltas are all smaller than the deltas registered for interventions (diagonal values). The two one-sided paired bootstrap tests indicate that 
the \genie scores across all features for paraphrased documents are statistically equivalent. This indicates \genie's robustness towards surface-level properties of texts.

\section{Evaluating Holistic Metrics}
Existing metrics that measure novelty, diversity and creativity are generally holistic, i.e., they generate 
one number for a given input,
making it unclear what aspects of novelty they are measuring. But \genie allows us to answer this question.
Specifically, we use a suite of baseline metrics to analyze their sensitivity in addition to \genie. We use the same setup as in \S\ref{sec:genie-causal-intervention}.

\subsection{Holistic Metrics} We measure the sensitivity of five holistic metrics that address novelty from different angles. 

\textbf{Cosine Distance:} We compute the average pairwise cosine distance of document embeddings using \texttt{text-embedding-vs-large} \cite{jiang2026artificial}.

\textbf{Compression Ratio (CR):} We calculate the compression ratio of the population with and without the target response using \texttt{gzip} to capture lexical diversity \cite{Shaib2024DetectionAM, shaib-etal-2025-standardizing}. 

\textbf{Creativity Index (Cr Idx):} We measure the semantic $n$-gram novelty for $n\geq L$ following \cite{lu2025ai} relative to a reference corpus.

\textbf{NoveltyBench Partition Model (NB):} We utilize the partition model from \cite{zhang2025noveltybench}, which clusters candidate responses based on their functional utility where utility is defined as how beneficial a response is in terms of information value.

\textbf{Litbench GenRM (LB):} We use the \texttt{Qwen-2.5-7b-Instruct} generative reward model fine-tuned on \citet{fein2025litbenchbenchmarkdatasetreliable} to calculate the average pairwise preference win-rates. 

All metrics are instantiated with the population used by \genie.
For instance, the corpora used by CreativityIndex  and the candidates considered by the partition model in NoveltyBench are both instantiated using our population.

\begin{table}[t!]
\small
\renewcommand{\arraystretch}{1}
\renewcommand{\tabcolsep}{1.0mm}
\centering
\begin{tabular}{lllllll}
\toprule
\multicolumn{7}{c}{\bf \genie v. Holistic Metrics Effect Size (Cohen's d)}\\
\multicolumn{7}{c}{*$p<0.008$, $^\dagger p<0.008$}   \\
\midrule
\multicolumn{1}{c}{Intervention} & \multicolumn{1}{c}{G} & \multicolumn{1}{c}{Cos.}  & \multicolumn{1}{c}{CR} & \multicolumn{1}{c}{Cr Idx} & \multicolumn{1}{c}{NB} & \multicolumn{1}{c}{LB}\\
\midrule

\Agent & 0.62* & 0.54* & \phantom{-}0.05 & 1.06* & 0.52* & -0.45* \\
\Perspective & 1.01* & 0.67* & -0.19* & 1.17* & 0.14 & -0.87* \\
\Plot & 0.71* & 0.22* & \phantom{-}0.01 & 0.93* & 0.30* & -0.33* \\
\Setting & 0.93* & 0.81* & \phantom{-}1.08* & 0.97* & 0.44* & -0.48* \\
\SocAtm & 0.47* & 0.24* & \phantom{-}0.23* & 1.00* & 0.16 & -0.42* \\
\Style & 0.38* & 0.68* & \phantom{-}0.07 & 1.43* & 0.07 & -1.53* \\

\midrule
Paraphrase & 0.03$^{\dagger}$ & 0.45$^{\dagger}$ & 1.25$^{\dagger}$ & 1.29 & 0.69 & -0.94 \\
\bottomrule
\end{tabular}
\vspace{-0.5em}
\caption{Effect sizes (Cohen's $d$) for feature interventions and paraphrases. \genie and Cosine Distance are sensitive to all interventions \textit{and} paraphrase robust. However, Cosine Distance weakly detects \Plot and \SocialAtmosphere. LitBench uniquely finds negative effect sizes, scoring intervened documents poorly. Refer to Table \ref{table:genie-sensitivty} for \genie's paraphrase effect sizes (note that the average is reported here).} \label{cohens-d-metrics}
\vspace{-1em}
\end{table}

\subsection{What Do Holistic Metrics Capture?}

We calculate both the effect sizes (Cohen's $d$) and statistical significance for scores found using each metric before and after intervention. 
This is shown in Table \ref{cohens-d-metrics}.
We note that the standard deviations vary across metrics, and report them in Table \ref{table:raw-mean-delta-stdev}.

\vspace{-0.3em}
\paragraph{Sensitivity}
Cosine Distance, Creativity Index and LitBench are sensitive to all feature interventions. Compression Ratio's exclusive sensitivity to \Setting interventions, which use highly descriptive terminology and out-of-distribution $n$-grams (e.g. \textit{rotating market pavilion on a lunar colony}; \textit{glass factory built inside a dormant volcanic caldera}), indicates that it does not capture most types of content novelty. NoveltyBench detects interventions moderately across most features, except \Perspective, \SocialAtmosphere and \Style. 
While most metrics perceive edited documents as more novel, LitBench scores them lower with negative effect sizes; we discuss this in a later paragraph.

\vspace{-0.3em}
\paragraph{Robustness}
The final row in Table \ref{cohens-d-metrics} shows Cohen's $d$ measured when using metrics on paraphrases. Values marked with $\dagger$ denote distributions that are statistically equivalent (we use two one-sided tests, see Appendix \ref{appendix:two-one-sided}).
Creativity Indiex, NoveltyBench and LitBench do not demonstrate robustness. Creativity Index appears to distinguish many types of interventions, but its lack of paraphrase robustness may indicate that it is overly sensitive to lexical novelty.

\begin{figure}
    \centering
    \includegraphics[width=.85\linewidth]{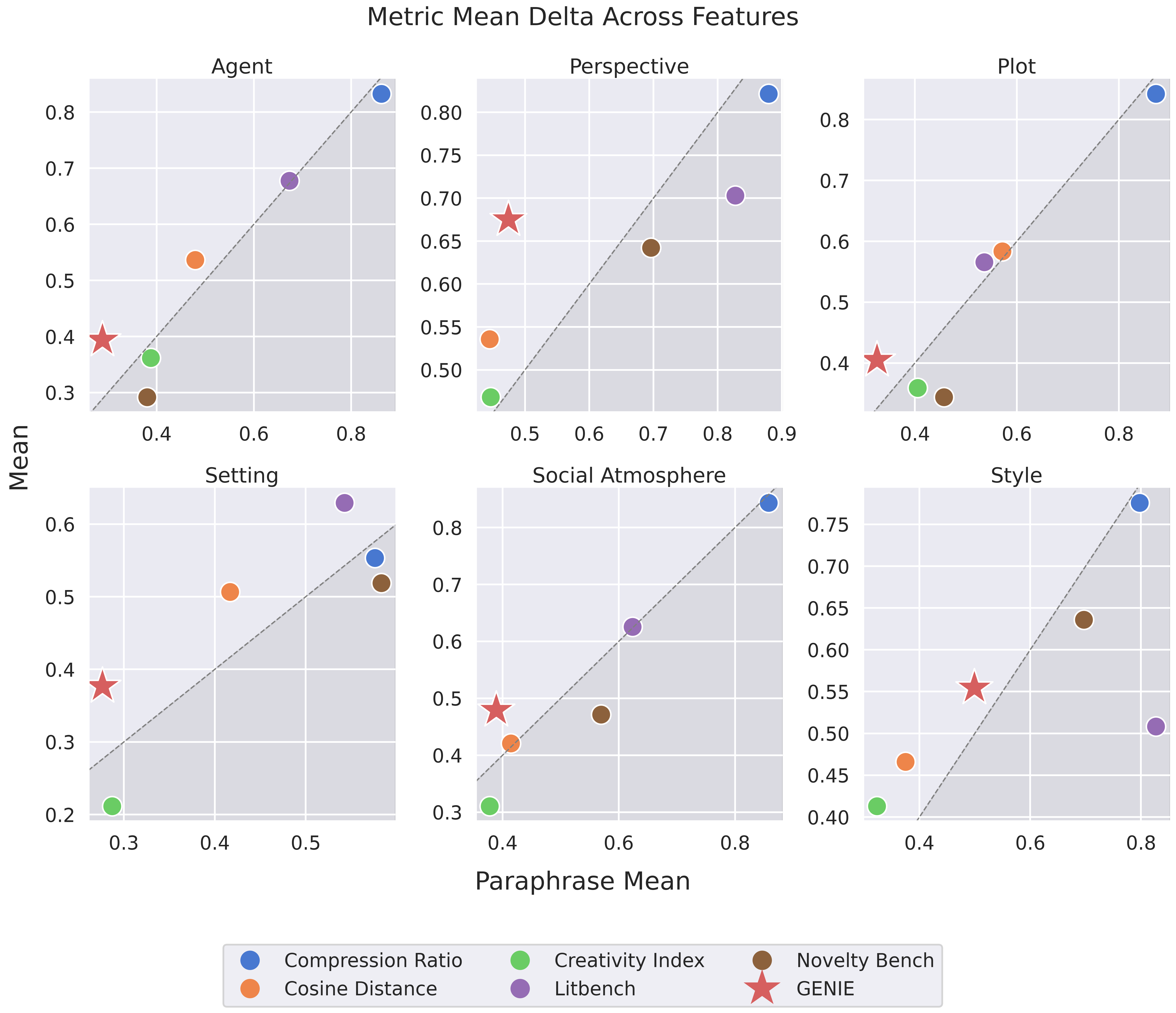}
    \caption{Normalized Mean Sensitivity and Paraphrase Absolute Deltas for each metric, per intervention. The shaded region represents higher average paraphrase sensitivity over interventions. \genie is consistently robust across all interventions. Holistic metrics, except Cosine Distance, are usually as equally or more sensitive to paraphrases than interventions.
    }
    \label{fig:sensitivty-paraphrase}
\end{figure}

Figure \ref{fig:sensitivty-paraphrase} shows the min-max normalized mean absolute deltas for interventions versus paraphrases for each metric and feature intervention. Metrics in the shaded region are not robust as they detect higher average paraphrase deltas than for feature interventions. \genie is the most consistently robust metric across all features. Compression Ratio and NoveltyBench consistently register a larger difference between paraphrased documents than intervened documents.

\vspace{-0.3em}
\paragraph{Aesthetic Quality}
\label{sec:litbench}
LitBench consistently registers negative effect sizes across all interventions and paraphrases. 
Upon close inspection, we see that
LitBench is capturing the degradation of a property consistent across edits. 
A qualitative analysis revealed that interventions, though accurate (Table \ref{table:intervention-accuracy}), are artificial and forced at times (e.g. \Perspective interventions include rewriting the story from \textit{first-person plural from the collective voice of the canned goods}).
Thus, we note that the palatability of edited documents may be degraded, and correlated with LitBench.
Although novelty and quality have a trade-off relation in model-generated responses \cite{West2025BaseMB}, they are independent axes of creativity \cite{boden1991creativemind}. Thus, metrics that capture aesthetic value are not suitable novelty metrics.

\vspace{-0.3em}
\paragraph{Functional Utility}
Since the NoveltyBench partition model was optimized for short-context tasks rather than  long-form creative writing, it is unclear which properties of a response determine its functional utility. However, the features it does not capture (\Perspective, \SocialAtmosphere and \Style)
are properties that are peripheral to the document (i.e. not explicitly stated in the document). Thus, we hypothesize that functional utility correlates with, and may be capturing, visible properties, lexical or semantic.

\paragraph{Spurious Correlations}
Table \ref{table:cosine-mean-score} shows the raw mean scores calculated by Cosine Distance and \genie per intervention. For certain interventions, such as \Plot, \genie appears to register a more dramatic change in novelty. Cosine Distance registers the same delta for \Perspective and \Style interventions ($0.05$), while \genie registers a very large delta for \Perspective, but not \Style. This suggests that Cosine Distance and \genie are capturing different properties (refer to Table \ref{table:cosine-corr} for a correlation study).

\begin{table}[t!]
\small
\renewcommand{\arraystretch}{1}
\renewcommand{\tabcolsep}{1.0mm}
\centering
\begin{tabular}{lcccc}
\toprule
\multicolumn{5}{c}{\bf Cosine Distance vs. \genie}  \\

\midrule
\multicolumn{1}{l}{Feature} & \multicolumn{2}{c}{Cos. Dist.} & \multicolumn{2}{c}{\genie}\\

\midrule
\multicolumn{1}{l}{} & \multicolumn{1}{c}{Orig} & \multicolumn{1}{c}{Edit} & \multicolumn{1}{c}{Orig} & \multicolumn{1}{c}{Edit}\\
\midrule

\Agent & 0.34 & 0.36 & 0.64 & 0.72\\
\Perspective & 0.32 & 0.37 & 0.37 & 0.80\\
\Plot & 0.34 & 0.34 & 0.67 & 0.72\\
\Setting & 0.33 & 0.36 & 0.61 & 0.75\\
\SocAtm & 0.33 & 0.34 & 0.68 & 0.74\\
\Style & 0.34 & 0.39 & 0.56 & 0.64\\
\bottomrule
\end{tabular}
\vspace{-0.5em}
\caption{The raw mean scores for Cosine Distance and \genie pre- and post- intervention.
} \label{table:cosine-mean-score}
\vspace{-1em}
\end{table}

\section{Effectiveness of Mitigation Methods}
The reduced diversity and creativity in model-generated responses \cite{kirk2024understanding, zhang2025noveltybench, jiang2026artificial} has motivated work that aims to mitigate this behavior and improve various dimensions of creativity. Some of these methods are evaluated using holistic metrics, such as pairwise cosine similarity and lexical metrics \cite{Zhang2025VerbalizedSH, peeperkorn2025mind}, as well as human and/or LLM judges \cite{huot2025agents, Chakrabarty2024CanAW, Pham2025FrankentextSR}. We use \genie to investigate the effectiveness of a subset of such methods.
Specifically, we 
measure the distributional differences observed pre- and post- mitigation. For the following experiments, we aggregate the \genie scores over all generations per prompt to form paired data.

\subsection{Target Methods}

We consider two broad categories of mitigation methods: alternate prompting and decoding based methods. 

We evaluate three prompting strategies: 
\textbf{(1) Verbalized Sampling} (VS) refers to generating \textit{n} samples using the target model $M$ with their associated probabilities \ramya{\cite{Zhang2025VerbalizedSH}}. \textbf{(2)} We use Agents' Room style prompting \cite{huot2025agents} to form detailed \textbf{Outlines} for each prompt and generate a document given the outline. This outlining requires the model to explicitly plan the Setting, Characters, Plot, etc., which closely follow the features of creative writing we use. \textbf{(3)} \textbf{Negative Sampling} (NS) is defined as generating a document $d$ using prompt $p$ and adding $d$ to the context and prompting the model to generate a response \textit{different} from $d$ \cite{zhang2025noveltybench}.

Finally, we explore \textbf{Conformative Decoding} (CD), a decoding strategy that blends base and instruct models at the token-level \cite{peeperkorn2025mind}. By using the weighted sum of token probabilities, the instruct model helps improve the quality while the base model introduces diverse content.\footnote{We weight base models higher than instruct models (0.6 and 0.4, respectively) to inspire more diverse outputs.}

Using the methods above, we generate responses to the prompts mentioned in \S\ref{sec:tasks} for the following models: \texttt{gemma-3-12b-it}, \texttt{gemma-3-27b-it}, \texttt{Llama-3.1-8B-Instruct}, \texttt{OLMo-3-7B-Instruct} and \texttt{OLMo-3-32B-Instruct}. Using these methods, we sample $5$ responses for each prompt, per model. We then calculate \genie to study the fine-grained improvements mitigation helps bring.

\subsection{Results}
\genie scores 
for each feature are aggregated across generations of documents per prompt, per model. For Negative Sampling, specifically, we pair the data at the instance level since target responses are paired prior to generating the data (adding the response to the context). We perform a paired bootstrap two-sided test between \genie scores for responses generated pre- and post- mitigation.
The mean \genie deltas are reported in in Table \ref{table:mitigation-effect-size}, with significance marked.

Outlining helps improve \Perspective and \Style, while Conformative Decoding improves \Setting. Negative Sampling has a significant improvement relative to the original counterpart. This is consistent with the findings of \citet{zhang2025noveltybench} regarding the effectiveness of this method. We note that despite significance, all mean deltas are small (e.g. relative to \genie detecting a delta of $0.35$ on \Perspective interventions in Figure \ref{fig:genie-all-scores}).

Although Verbalized Sampling \textit{appears} to be the most effective mitigation method, Table \ref{table:mitigation-effect-size} shows that responses generated by this method are disproportionately shorter relative to the target and population set. These lengths ($\sim$100 words) are similar to flash fiction with $2$-$3$ sentences, rather than a fully formed story (e.g. Appendix \ref{example:vs-docs}). Thus, comparing the novelty of these documents with respect to a population that is from a different distribution
is not representative. This highlights the importance of using an appropriate population when computing novelty. 

This analysis illustrates what fine-grained evaluation enables, rather than a single aggregate score, \genie attributes each method's effect to specific features.

\begin{table}[t!]
\small
\renewcommand{\arraystretch}{1}
\renewcommand{\tabcolsep}{1.0mm}
\centering
\begin{tabular}{lcccc}
\toprule
\multicolumn{5}{c}{\bf Mean \genie Delta (*p<0.008)} \\
\midrule
\multicolumn{1}{c}{Feature} &\multicolumn{1}{c}{\bf VS} &\multicolumn{1}{c}{\bf Outline} &\multicolumn{1}{c}{\bf NS} &\multicolumn{1}{c}{\bf CD}\\
\midrule
\Agent & 0.03* & 0.02\phantom{*} & 0.02* & 0.01\phantom{*}\\
\Perspective & 0.01\phantom{*} & 0.03* & 0.01\phantom{*} & 0.01\phantom{*}\\
\Plot & 0.02* & 0.00\phantom{*} & 0.02* & 0.01\phantom{*}\\
\Setting & 0.04* & 0.00\phantom{*} & 0.03* & 0.02*\\
\SocAtm & 0.03* & 0.00\phantom{*} & 0.01* & 0.01\phantom{*}\\
\Style & 0.02* & 0.02* & 0.02* & 0.00\phantom{*}\\
\midrule
Avg. num words & \cellcolor{redhl!40}98 & 513 & 510 & 398 \\
\midrule
Avg. Coherence & 3.45 &3.35 & 3.32 & 3.19\\
\bottomrule
\end{tabular}
\vspace{-0.5em}
\caption{Mean \genie deltas observed when using mitigation methods. Significance is marked with (*). Conformative Decoding helps improve \Setting and Outlinging improves \Perspective and \Style. The average number of words in Verbalized Sampling scores are not representative as responses are disproportionately small relative to the population (488) and target (513) set.  Average Coherence computed using ArmoRM (out of 5) is moderate across all methods.
} \label{table:mitigation-effect-size}
\vspace{-1em}
\end{table}

\section{Conclusion}
We present \genie, a fine-grained evaluation metric to measure the \textit{novelty} of LLM-generated responses with respect to an explicitly defined population. \genie discovers and grounds task-specific \textit{features} in responses using answers to prompt-specific questions. We instantiate \genie in the creative writing domain as a testbed and show that \genie is sensitive to features when independently intervened upon and robust towards lexical edits (i.e. paraphrases). Using \genie, we study the sensitivity and robustness of existing holistic creativity, diversity and novelty and find that they either capture extraneous properties or are lexically sensitive. Finally, we evaluate the effectiveness of mitigation methods using a fine-grained lens, revealing the strengths of alternate prompting and decoding strategies.

\section*{Limitations} 

\paragraph{Computational Cost} The \genie pipeline is moderately resource-intensive. We use proprietary models for feature discovery, question generation, question-to-feature mapping, alternate feature generation, document editing, and paraphrasing, and \texttt{Qwen2.5-32B-Instruct} for question answering and similarity judgments. Using \genie for other domains requires generating new populations because \genie is heavily task-oriented. That said, measuring novelty is really difficult, and we view the compute cost worth the cost of going beyond the coarse, surface-level proxies to explainable, feature-grounded evaluation (see Appendix \ref{appendix:cost} for more details).

\paragraph{Domain Scope} We instantiate \genie in the creative writing domain. The pipeline is, by design, generalizable to any text-based task but we have not run extensive experiments measuring how well existing metrics capture novelty outside creative writing. Thus, this pipeline has not been validated for use in domains such as scientific writing, code or argumentative essays.

\section*{Acknowledgments}

We thank Kathryn Kazanas for helping with preliminary annotations, Chantal Shaib for early guidance and the TAUR Lab for feedback on the project. This work was partially supported by NSF grants IIS-2107524, IIS-2145479, IIS-2145280, and IIS-2433071, as well as the Sloan Foundation. Compute for this project was partially supplied by the Machine Learning Lab at UT Austin and the NSF AI Institute for Foundations of Machine Learning (IFML).

\bibliography{custom}

@article{Mesnard2024GemmaOM,
  title={Gemma: Open Models Based on Gemini Research and Technology},
  author={Thomas Mesnard and Cassidy Hardin and Robert Dadashi and Surya Bhupatiraju and Shreya Pathak and L. Sifre and Morgane Rivi{\`e}re and Mihir Kale and J Christopher Love and Pouya Dehghani Tafti and L'eonard Hussenot and Aakanksha Chowdhery and Adam Roberts and Aditya Barua and Alex Botev and Alex Castro-Ros and Ambrose Slone and Am'elie H'eliou and Andrea Tacchetti and Anna Bulanova and Antonia Paterson and Beth Tsai and Bobak Shahriari and Charline Le Lan and Christopher A. Choquette-Choo and Cl{\'e}-ment Crepy and Daniel Cer and Daphne Ippolito and David Reid and Elena Buchatskaya and Eric Ni and Eric Noland and Geng Yan and George Tucker and George-Christian Muraru and Grigory Rozhdestvenskiy and Henryk Michalewski and Ian Tenney and Ivan Grishchenko and Jacob Austin and James Keeling and Jane Labanowski and Jean-Baptiste Lespiau and Jeff Stanway and Jenny Brennan and Jeremy Chen and Johan Ferret and Justin Chiu and Justin Mao-Jones and Kather-ine Lee and Kathy Yu and Katie Millican and Lars Lowe Sjoesund and Lisa Lee and Lucas Dixon and Machel Reid and Maciej Mikuła and Mateo Wirth and Michael Sharman and Nikolai Chinaev and Nithum Thain and Olivier Bachem and Os-car Chang and Oscar Wahltinez and Paige Bailey and Paul Michel and Petko Yotov and Pier Giuseppe Sessa and Rahma Chaabouni and Ramona Comanescu and Reena Jana and Rohan Anil and Ross Mcilroy and Ruibo Liu and Ryan Mullins and Samuel L. Smith and Sebastian Borgeaud and Sertan Girgin and Sholto Douglas and Shree Pandya and Siamak Shakeri and Soham De and Ted Klimenko and Tom Hennigan and Vladimir Feinberg and Wojciech Stokowiec and Yu-Hui Chen and Zafarali Ahmed and Zhitao Gong and Tris Warkentin and Ludovic Peran and Minh Giang and Cl{\'e}ment Farabet and Oriol Vinyals and Jeffrey Dean and Koray Kavukcuoglu and Demis Hassabis and Zoubin Ghahramani and Douglas Eck and Joelle Barral and Fernando Pereira and Eli Collins and Armand Joulin and Noah Fiedel and Evan Senter and Alek Andreev and Kathleen Kenealy},
  journal={ArXiv},
  year={2024},
  volume={abs/2403.08295},
  url={https://api.semanticscholar.org/CorpusID:268379206}
}

@inproceedings{shaib-etal-2025-standardizing,
    title = "Standardizing the Measurement of Text Diversity: A Tool and Comparative Analysis",
    author = "Shaib, Chantal  and
      Govindarajan, Venkata S  and
      Barrow, Joe  and
      Sun, Jiuding  and
      Siu, Alexa  and
      Wallace, Byron C  and
      Nenkova, Ani",
    editor = "Liu, Xuebo  and
      Purwarianti, Ayu",
    booktitle = "Proceedings of The 14th International Joint Conference on Natural Language Processing and The 4th Conference of the Asia-Pacific Chapter of the Association for Computational Linguistics: System Demonstrations",
    month = dec,
    year = "2025",
    address = "Mumbai, India",
    publisher = "Association for Computational Linguistics",
    url = "https://aclanthology.org/2025.ijcnlp-demo.5/",
    doi = "10.18653/v1/2025.ijcnlp-demo.5",
    pages = "36--46",
    ISBN = "979-8-89176-301-2"
}

@inproceedings{Yang2025Qwen3TR,
  title={Qwen3 Technical Report},
  author={An Yang and Anfeng Li and Baosong Yang and Beichen Zhang and Binyuan Hui and Bo Zheng and Bowen Yu and Chang Gao and Chengen Huang and Chenxu Lv and Chujie Zheng and Dayiheng Liu and Fan Zhou and Fei Huang and Feng Hu and Hao Ge and Haoran Wei and Huan Lin and Jialong Tang and Jian Yang and Jianhong Tu and Jianwei Zhang and Jianxin Yang and Jiaxin Yang and Jingren Zhou and Jingren Zhou and Junyan Lin and Kai Dang and Keqin Bao and Ke‐Pei Yang and Le Yu and Li-Chun Deng and Mei Li and Min Xue and Mingze Li and Pei Zhang and Peng Wang and Qin Zhu and Rui Men and Ruize Gao and Shi-Qiang Liu and Shuang Luo and Tianhao Li and Tianyi Tang and Wenbiao Yin and Xingzhang Ren and Xinyu Wang and Xinyu Zhang and Xuancheng Ren and Yang Fan and Yang Su and Yi-Chao Zhang and Yinger Zhang and Yu Wan and Yuqiong Liu and Zekun Wang and Zeyu Cui and Zhenru Zhang and Zhipeng Zhou and Zihan Qiu},
  year={2025},
  url={https://api.semanticscholar.org/CorpusID:278602855}
}

@misc{olmo2026olmo3,
      title={Olmo 3}, 
      author={Allyson Ettinger and Amanda Bertsch and Bailey Kuehl and David Graham and David Heineman and Dirk Groeneveld and Faeze Brahman and Finbarr Timbers and Hamish Ivison and Jacob Morrison and Jake Poznanski and Kyle Lo and Luca Soldaini and Matt Jordan and Mayee Chen and Michael Noukhovitch and Nathan Lambert and Pete Walsh and Pradeep Dasigi and Robert Berry and Saumya Malik and Saurabh Shah and Scott Geng and Shane Arora and Shashank Gupta and Taira Anderson and Teng Xiao and Tyler Murray and Tyler Romero and Victoria Graf and Akari Asai and Akshita Bhagia and Alexander Wettig and Alisa Liu and Aman Rangapur and Chloe Anastasiades and Costa Huang and Dustin Schwenk and Harsh Trivedi and Ian Magnusson and Jaron Lochner and Jiacheng Liu and Lester James V. Miranda and Maarten Sap and Malia Morgan and Michael Schmitz and Michal Guerquin and Michael Wilson and Regan Huff and Ronan Le Bras and Rui Xin and Rulin Shao and Sam Skjonsberg and Shannon Zejiang Shen and Shuyue Stella Li and Tucker Wilde and Valentina Pyatkin and Will Merrill and Yapei Chang and Yuling Gu and Zhiyuan Zeng and Ashish Sabharwal and Luke Zettlemoyer and Pang Wei Koh and Ali Farhadi and Noah A. Smith and Hannaneh Hajishirzi},
      year={2026},
      eprint={2512.13961},
      archivePrefix={arXiv},
      primaryClass={cs.CL},
      url={https://arxiv.org/abs/2512.13961}, 
}

@inproceedings{Shaib2024DetectionAM,
  title={{Detection and Measurement of Syntactic Templates in Generated Text}},
  author={Chantal Shaib and Yanai Elazar and Junyi Jessy Li and Byron C. Wallace},
  booktitle={Conference on Empirical Methods in Natural Language Processing},
  year={2024},
  url={https://api.semanticscholar.org/CorpusID:270869797}
}

@inproceedings{Russell2026StoryScopeII,
  title={{StoryScope: Investigating Idiosyncrasies in AI Fiction}},
  author={Jenna Russell and Rishanth Rajendhran and Chau Minh Pham and Mohit Iyyer and John Wieting},
  year={2026},
  booktitle={Third Conference on Language Modeling},
  url={https://api.semanticscholar.org/CorpusID:287122180}
}

@inproceedings{
cao2026evaluating,
title={{Evaluating Text Creativity across Diverse Domains: a Dataset and Large Language Model Evaluator}},
author={Qian Cao and Xiting Wang and Yuzhuo Yuan and Yahui Liu and Fang Luo and Ruihua Song},
booktitle={The Fourteenth International Conference on Learning Representations},
year={2026},
url={https://openreview.net/forum?id=WHrmlyQEM7}
}

@article{arsteinInterCoderAgrCL,
    author = {Artstein, Ron and Poesio, Massimo},
    title = {{Inter-Coder Agreement for Computational Linguistics}},
    journal = {Computational Linguistics},
    volume = {34},
    number = {4},
    pages = {555-596},
    year = {2008},
    month = {12},
    issn = {0891-2017},
    doi = {10.1162/coli.07-034-R2},
    url = {https://doi.org/10.1162/coli.07-034-R2},
    eprint = {https://direct.mit.edu/coli/article-pdf/34/4/555/1808947/coli.07-034-r2.pdf},
}

@inproceedings{Chakrabarty2024CanAW,
author = {Chakrabarty, Tuhin and Laban, Philippe and Wu, Chien-Sheng},
title = {{Can AI writing be salvaged? Mitigating Idiosyncrasies and Improving Human-AI Alignment in the Writing Process through Edits}},
year = {2025},
isbn = {9798400713941},
publisher = {Association for Computing Machinery},
address = {New York, NY, USA},
url = {https://doi.org/10.1145/3706598.3713559},
doi = {10.1145/3706598.3713559},
booktitle = {Proceedings of the 2025 CHI Conference on Human Factors in Computing Systems},
articleno = {1210},
numpages = {33},
location = {
},
series = {CHI '25}
}

@inproceedings{Pham2025FrankentextSR,
    title = "Frankentext: Stitching random text fragments into long-form narratives",
    author = "Pham, Chau Minh  and
      Russell, Jenna  and
      Pham, Dzung  and
      Iyyer, Mohit",
    editor = "Liakata, Maria  and
      Moreira, Viviane P.  and
      Zhang, Jiajun  and
      Jurgens, David",
    booktitle = "Proceedings of the 64th Annual Meeting of the {A}ssociation for {C}omputational {L}inguistics (Volume 1: Long Papers)",
    month = jul,
    year = "2026",
    address = "San Diego, California, United States",
    publisher = "Association for Computational Linguistics",
    url = "https://aclanthology.org/2026.acl-long.1457/",
    doi = "10.18653/v1/2026.acl-long.1457",
    pages = "31586--31625",
    ISBN = "979-8-89176-390-6"
}

@inproceedings{
kirk2024understanding,
title={{Understanding the Effects of {RLHF} on {LLM} Generalisation and Diversity}},
author={Robert Kirk and Ishita Mediratta and Christoforos Nalmpantis and Jelena Luketina and Eric Hambro and Edward Grefenstette and Roberta Raileanu},
booktitle={The Twelfth International Conference on Learning Representations},
year={2024},
url={https://openreview.net/forum?id=PXD3FAVHJT}
}

@inproceedings{le-bronnec-etal-2024-exploring,
    title = {{Exploring Precision and Recall to assess the quality and diversity of {LLM}s}},
    author = "Le Bronnec, Florian  and
      Verine, Alexandre  and
      Negrevergne, Benjamin  and
      Chevaleyre, Yann  and
      Allauzen, Alexandre",
    editor = "Ku, Lun-Wei  and
      Martins, Andre  and
      Srikumar, Vivek",
    booktitle = "Proceedings of the 62nd Annual Meeting of the Association for Computational Linguistics (Volume 1: Long Papers)",
    month = aug,
    year = "2024",
    address = "Bangkok, Thailand",
    publisher = "Association for Computational Linguistics",
    url = "https://aclanthology.org/2024.acl-long.616/",
    doi = "10.18653/v1/2024.acl-long.616",
    pages = "11418--11441"
}

@inproceedings{
huot2025agents,
title={{Agents' Room:  Narrative Generation through Multi-step Collaboration}},
author={Fantine Huot and Reinald Kim Amplayo and Jennimaria Palomaki and Alice Shoshana Jakobovits and Elizabeth Clark and Mirella Lapata},
booktitle={The Thirteenth International Conference on Learning Representations},
year={2025},
url={https://openreview.net/forum?id=HfWcFs7XLR}
}

@inproceedings{fein2025litbenchbenchmarkdatasetreliable,
    title = {{{L}it{B}ench: A Benchmark and Dataset for Reliable Evaluation of Creative Writing}},
    author = "Fein, Daniel  and
      Russo, Sebastian  and
      Xiang, Violet  and
      Jolly, Kabir  and
      Rafailov, Rafael  and
      Haber, Nick",
    editor = "Demberg, Vera  and
      Inui, Kentaro  and
      Marquez, Llu{\'i}s",
    booktitle = "Proceedings of the 19th Conference of the {E}uropean Chapter of the {A}ssociation for {C}omputational {L}inguistics (Volume 1: Long Papers)",
    month = mar,
    year = "2026",
    address = "Rabat, Morocco",
    publisher = "Association for Computational Linguistics",
    url = "https://aclanthology.org/2026.eacl-long.362/",
    doi = "10.18653/v1/2026.eacl-long.362",
    pages = "7740--7755",
    ISBN = "979-8-89176-380-7"
}

@article{Laban2026LLMsCY,
  title={{LLMs Corrupt Your Documents When You Delegate}},
  author={Philippe Laban and Tobias Schnabel and Jennifer Neville},
  year={2026},
  journal={arXiv preprint},
  url={https://api.semanticscholar.org/CorpusID:287607479}
}

@inproceedings{Chakrabarty2023ArtOA,
author = {Chakrabarty, Tuhin and Laban, Philippe and Agarwal, Divyansh and Muresan, Smaranda and Wu, Chien-Sheng},
title = {{Art or Artifice? Large Language Models and the False Promise of Creativity}},
year = {2024},
isbn = {9798400703300},
publisher = {Association for Computing Machinery},
address = {New York, NY, USA},
url = {https://doi.org/10.1145/3613904.3642731},
doi = {10.1145/3613904.3642731},
booktitle = {Proceedings of the 2024 CHI Conference on Human Factors in Computing Systems},
articleno = {30},
numpages = {34},
location = {Honolulu, HI, USA},
series = {CHI '24}
}

@inproceedings{
jiang2026artificial,
title={{Artificial Hivemind: The Open-Ended Homogeneity of Language Models (and Beyond)}},
author={Liwei Jiang and Yuanjun Chai and Margaret Li and Mickel Liu and Raymond Fok and Nouha Dziri and Yulia Tsvetkov and Maarten Sap and Yejin Choi},
booktitle={The Thirty-ninth Annual Conference on Neural Information Processing Systems Datasets and Benchmarks Track},
year={2026},
url={https://openreview.net/forum?id=saDOrrnNTz}
}

@inproceedings{
chakrabarty2025aislop,
title={{{AI}-Slop to {AI}-Polish? Aligning Language Models through Edit-Based Writing Rewards and Test-time computation}},
author={Tuhin Chakrabarty and Philippe Laban and Chien-Sheng Wu},
booktitle={Second Conference on Language Modeling},
year={2025},
url={https://openreview.net/forum?id=jeDYcjuZIV}
}

@inproceedings{
zhang2025noveltybench,
title={{NoveltyBench: Evaluating Creativity and Diversity in Language Models}},
author={Yiming Zhang and Harshita Diddee and Susan Holm and Hanchen Liu and Xinyue Liu and Vinay Samuel and Barry Wang and Daphne Ippolito},
booktitle={Second Conference on Language Modeling},
year={2025},
url={https://openreview.net/forum?id=XZm1ekzERf}
}

@inproceedings{
lu2025ai,
title={{{AI} as Humanity{\textquoteright}s Salieri: Quantifying Linguistic Creativity of Language Models via Systematic Attribution of Machine Text against Web Text}},
author={Ximing Lu and Melanie Sclar and Skyler Hallinan and Niloofar Mireshghallah and Jiacheng Liu and Seungju Han and Allyson Ettinger and Liwei Jiang and Khyathi Chandu and Nouha Dziri and Yejin Choi},
booktitle={The Thirteenth International Conference on Learning Representations},
year={2025},
url={https://openreview.net/forum?id=ilOEOIqolQ}
}

@inproceedings{ismayilzada2025creative,
    title = {{Creative Preference Optimization}},
    author = "Ismayilzada, Mete  and
      Laverghetta Jr., Antonio  and
      Luchini, Simone A.  and
      Patel, Reet  and
      Bosselut, Antoine  and
      Plas, Lonneke Van Der  and
      Beaty, Roger E.",
    editor = "Christodoulopoulos, Christos  and
      Chakraborty, Tanmoy  and
      Rose, Carolyn  and
      Peng, Violet",
    booktitle = "Findings of the Association for Computational Linguistics: EMNLP 2025",
    month = nov,
    year = "2025",
    address = "Suzhou, China",
    publisher = "Association for Computational Linguistics",
    url = "https://aclanthology.org/2025.findings-emnlp.509/",
    doi = "10.18653/v1/2025.findings-emnlp.509",
    pages = "9580--9609",
    ISBN = "979-8-89176-335-7"
}

@book{boden1991creativemind,
author = {Boden, Margaret A.},
title = {{The Creative Mind: Myths and Mechanisms}},
year = {1991},
isbn = {0465014526},
publisher = {Basic Books, Inc.},
address = {USA}
}

@article{Wenger2025WereDW,
  title={{We're Different, We're the Same: Creative Homogeneity Across LLMs}},
  author={Emily Wenger and Yoed N. Kenett},
  journal={ArXiv},
  year={2025},
  volume={abs/2501.19361},
  url={https://api.semanticscholar.org/CorpusID:276079902}
}

@inproceedings{
Lanchantin2025DiversePO,
title={{Diverse Preference Learning for Capabilities and Alignment}},
author={Stewart Slocum and Asher Parker-Sartori and Dylan Hadfield-Menell},
booktitle={The Thirteenth International Conference on Learning Representations},
year={2025},
url={https://openreview.net/forum?id=pOq9vDIYev}
}

@inproceedings{
West2025BaseMB,
title={{Base Models Beat Aligned Models at Randomness and Creativity}},
author={Peter West and Christopher Potts},
booktitle={Second Conference on Language Modeling},
year={2025},
url={https://openreview.net/forum?id=vqN8uom4A1}
}

@inproceedings{Hamilton2025NarraBenchAC,
    title = {{{N}arra{B}ench: A Comprehensive Framework for Narrative Benchmarking}},
    author = "Hamilton, Sil  and
      Wilkens, Matthew  and
      Piper, Andrew",
    editor = "Demberg, Vera  and
      Inui, Kentaro  and
      Marquez, Llu{\'i}s",
    booktitle = "Proceedings of the 19th Conference of the {E}uropean Chapter of the {A}ssociation for {C}omputational {L}inguistics (Volume 1: Long Papers)",
    month = mar,
    year = "2026",
    address = "Rabat, Morocco",
    publisher = "Association for Computational Linguistics",
    url = "https://aclanthology.org/2026.eacl-long.176/",
    doi = "10.18653/v1/2026.eacl-long.176",
    pages = "3786--3801",
    ISBN = "979-8-89176-380-7"
}

@inproceedings{Hamilton2024DetectingMC,
    title = {{Detecting Mode Collapse in Language Models via Narration}},
    author = "Hamilton, Sil",
    editor = "Miceli-Barone, Antonio Valerio  and
      Barez, Fazl  and
      Cohen, Shay B.  and
      Voita, Elena  and
      Germann, Ulrich  and
      Lukasik, Michal",
    booktitle = "Proceedings of the First edition of the Workshop on the Scaling Behavior of Large Language Models (SCALE-LLM 2024)",
    month = mar,
    year = "2024",
    address = "St. Julian{'}s, Malta",
    publisher = "Association for Computational Linguistics",
    url = "https://aclanthology.org/2024.scalellm-1.5/",
    doi = "10.18653/v1/2024.scalellm-1.5",
    pages = "65--72"
}

@inproceedings{
Zhang2025VerbalizedSH,
title={Verbalized Sampling: How to Mitigate Mode Collapse and Unlock {LLM} Diversity},
author={Jiayi Zhang and Simon Yu and Derek Chong and Anthony Sicilia and Michael Tomz and Christopher D Manning and Weiyan Shi},
booktitle={Forty-third International Conference on Machine Learning},
year={2026},
url={https://openreview.net/forum?id=zloIrd77G5}
}

@inproceedings{
padmakumar2026measuring,
title={{Measuring {LLM} Novelty As The Frontier Of Original And High-Quality Output}},
author={Vishakh Padmakumar and Chen Yueh-Han and Jane Pan and Valerie Chen and He He},
booktitle={The Fourteenth International Conference on Learning Representations},
year={2026},
url={https://openreview.net/forum?id=i7QNKZioN6}
}

@inproceedings{
padmakumar2024does,
title={{Does Writing with Language Models Reduce Content Diversity?}},
author={Vishakh Padmakumar and He He},
booktitle={The Twelfth International Conference on Learning Representations},
year={2024},
url={https://openreview.net/forum?id=Feiz5HtCD0}
}

@article{franceschelli2024creativity,
author = {Franceschelli, Giorgio and Musolesi, Mirco},
title = {{On the Creativity of Large Language Models}},
year = {2024},
issue_date = {Jun 2025},
publisher = {Springer-Verlag},
address = {Berlin, Heidelberg},
volume = {40},
number = {5},
issn = {0951-5666},
url = {https://doi.org/10.1007/s00146-024-02127-3},
doi = {10.1007/s00146-024-02127-3},
journal = {AI Soc.},
month = nov,
pages = {3785–3795},
numpages = {11}
}

@article{schapiro2025combinatorial,
  title={{Combinatorial Creativity: A New Frontier in Generalization Abilities}},
  author={Schapiro, Samuel and Shashidhar, Sumuk and Gladstone, Alexi and Black, Jonah and Moon, Royce and Hakkani-Tur, Dilek and Varshney, Lav R},
  journal={arXiv preprint arXiv:2509.21043},
  year={2025}
}

@inproceedings{
nagarajan2025roll,
title={{Roll the dice \& look before you leap:
Going beyond the creative limits of next-token prediction}},
author={Vaishnavh Nagarajan and Chen Henry Wu and Charles Ding and Aditi Raghunathan},
booktitle={Forty-second International Conference on Machine Learning},
year={2025},
url={https://openreview.net/forum?id=Hi0SyHMmkd}
}

@article{wadhwa2026create,
  title={{CREATE: Testing LLMs for Associative Creativity
}},
  author={Wadhwa, Manya and Roy, Tiasa Singha and Lederman, Harvey and Li, Junyi Jessy and Durrett, Greg},
  journal={arXiv preprint arXiv:2603.09970},
  year={2026}
}

@article{
xu2025echoes,
author = {Weijia Xu  and Nebojsa Jojic  and Sudha Rao  and Chris Brockett  and Bill Dolan },
title = {{Echoes in AI: Quantifying lack of plot diversity in LLM outputs}},
journal = {Proceedings of the National Academy of Sciences},
volume = {122},
number = {35},
pages = {e2504966122},
year = {2025},
doi = {10.1073/pnas.2504966122},
URL = {https://www.pnas.org/doi/abs/10.1073/pnas.2504966122},
eprint = {https://www.pnas.org/doi/pdf/10.1073/pnas.2504966122}}

@inproceedings{merrill-etal-2024-evaluating,
    title = {{Evaluating $n$-Gram Novelty of Language Models Using Rusty-{DAWG}}},
    author = "Merrill, William  and
      Smith, Noah A.  and
      Elazar, Yanai",
    editor = "Al-Onaizan, Yaser  and
      Bansal, Mohit  and
      Chen, Yun-Nung",
    booktitle = "Proceedings of the 2024 Conference on Empirical Methods in Natural Language Processing",
    month = nov,
    year = "2024",
    address = "Miami, Florida, USA",
    publisher = "Association for Computational Linguistics",
    url = "https://aclanthology.org/2024.emnlp-main.800/",
    doi = "10.18653/v1/2024.emnlp-main.800",
    pages = "14459--14473"
}

@inproceedings{
namuduri2025qudsim,
title={{{QUD}sim: Quantifying Discourse Similarities in {LLM}-Generated Text}},
author={Ramya Namuduri and Yating Wu and Anshun Asher Zheng and Manya Wadhwa and Greg Durrett and Junyi Jessy Li},
booktitle={Second Conference on Language Modeling},
year={2025},
url={https://openreview.net/forum?id=zFz1BJu211}
}

@inproceedings{fan2018hierarchical,
  title={{Hierarchical Neural Story Generation}},
  author={Fan, Angela and Lewis, Mike and Dauphin, Yann},
  booktitle={Proceedings of the 56th Annual Meeting of the Association for Computational Linguistics (Volume 1: Long Papers)},
  pages={889--898},
  year={2018}
}

@inproceedings{ArmoRM,
      title={{Interpretable Preferences via Multi-Objective Reward Modeling and Mixture-of-Experts}}, 
      author={Haoxiang Wang and Wei Xiong and Tengyang Xie and Han Zhao and Tong Zhang},
      booktitle={EMNLP},
      year={2024}
}

@article{Kamath2025Gemma3T,
  title={{Gemma 3 Technical Report}},
  author={Aishwarya Kamath and Johan Ferret and Shreya Pathak and Nino Vieillard and Ramona Merhej and Sarah Perrin and Tatiana Matejovicova and Alexandre Ram'e and Morgane Rivi{\`e}re and Louis Rouillard and Thomas Mesnard and Geoffrey Cideron and Jean-Bastien Grill and Sabela Ramos and Edouard Yvinec and Michelle Casbon and Etienne Pot and Ivo Penchev and Gael Liu and Francesco Visin and Kathleen Kenealy and Lucas Beyer and Xiaohai Zhai and Anton Tsitsulin and R{\'o}bert Istvan Busa-Fekete and Alex Feng and Noveen Sachdeva and Benjamin Coleman and Yi Gao and Basil Mustafa and Iain Barr and Emilio Parisotto and David Tian and Matan Eyal and Colin Cherry and Jan-Thorsten Peter and Danila Sinopalnikov and Surya Bhupatiraju and Rishabh Agarwal and Mehran Kazemi and Dan Malkin and Ravin Kumar and David Vilar and Idan Brusilovsky and Jiaming Luo and Andreas Steiner and Abe Friesen and Abhanshu Sharma and Abheesht Sharma and Adi Mayrav Gilady and Adrian Goedeckemeyer and Alaa Saade and Alexander Kolesnikov and Alexei Bendebury and Alvin Abdagic and Amit Vadi and Andr'as Gyorgy and Andr{\'e} Susano Pinto and Anil Das and Ankur Bapna and Antoine Miech and Antoine Yang and Antonia Paterson and Ashish Shenoy and Ayan Chakrabarti and Bilal Piot and Boxi Wu and Bobak Shahriari and Bryce Petrini and Charlie Chen and Charline Le Lan and Christopher A. Choquette-Choo and Cj Carey and Cormac Brick and Daniel Deutsch and Danielle Eisenbud and Dee Cattle and Derek Zhiyuan Cheng and Dimitris Paparas and Divyashree Shivakumar Sreepathihalli and Doug Reid and Dustin Tran and Dustin Zelle and Eric Noland and Erwin Huizenga and Eugene Kharitonov and Frederick Liu and Gagik Amirkhanyan and Glenn Cameron and Hadi Hashemi and Hanna Klimczak-Pluci'nska and Harman Singh and Harsh Mehta and Harshal Tushar Lehri and Hussein Hazimeh and Ian Ballantyne and Idan Szpektor and Ivan Nardini and Jean Pouget-Abadie and Jetha Chan and Joe Stanton and J. Michael Wieting and Jonathan Lai and Jordi Orbay and Joe Fernandez and Joshua Newlan and Junsong Ji and Jyotinder Singh and Kat Black and Kathy Yu and Kevin Hui and Kiran Vodrahalli and Klaus Greff and Linhai Qiu and Marcella Valentine and Marina Coelho and Marvin Ritter and Matt Hoffman and Matthew Watson and Mayank Chaturvedi and Michael Moynihan and Min Ma and Nabila Babar and Natasha Noy and Nathan Byrd and Nick Roy and Nikola Momchev and Nilay Chauhan and Oskar Bunyan and Pankil Botarda and Paul Caron and Paul Kishan Rubenstein and Phil Culliton and Philipp Schmid and Pier Giuseppe Sessa and Ping-mei Xu and Piotr Stańczyk and Pouya Dehghani Tafti and Rakesh Shivanna and Renjie Wu and Renke Pan and Reza Ardeshir Rokni and Rob Willoughby and Rohith Vallu and Ryan Mullins and Sammy Jerome and Sara Smoot and Sertan Girgin and Shariq Iqbal and Shashir Reddy and Shruti Sheth and Siim P{\~o}der and Sijal Bhatnagar and Sindhu Raghuram Panyam and Sivan Eiger and Susan Zhang and Tianqi Liu and Trevor Yacovone and Tyler Liechty and Uday Kalra and Utku Evci and Vedant Misra and Vincent Roseberry and Vladimir Feinberg and Vlad Kolesnikov and Woohyun Han and Woosuk Kwon and Xi Chen and Yinlam Chow and Yuvein Zhu and Zichuan Wei and Zoltan Egyed and Victor Cotruta and Minh Giang and Phoebe Kirk and Anand Rao and Jessica Lo and Erica Moreira and Luiz Gustavo Martins and Omar Sanseviero and Lucas Gonzalez and Zach Gleicher and Tris Warkentin and Vahab S. Mirrokni and Evan Senter and Eli Collins and Joelle Barral and Zoubin Ghahramani and Raia Hadsell and Yossi Matias and D. Sculley and Slav Petrov and Noah Fiedel and Noam Shazeer and Oriol Vinyals and Jeffrey Dean and Demis Hassabis and Koray Kavukcuoglu and Cl{\'e}ment Farabet and Elena Buchatskaya and Jean-Baptiste Alayrac and Rohan Anil and Dmitry Lepikhin and Sebastian Borgeaud and Olivier Bachem and Armand Joulin and Alek Andreev and Cassidy Hardin and Robert Dadashi and L'eonard Hussenot},
  journal={ArXiv},
  year={2025},
  volume={abs/2503.19786},
  url={https://api.semanticscholar.org/CorpusID:277313563}
}

@article{peeperkorn2025mind,
  publtype={informal},
  author={Max Peeperkorn and Tom Kouwenhoven and Dan Brown and Anna Jordanous},
  title={{Mind the Gap: Conformative Decoding to Improve Output Diversity of Instruction-Tuned Large Language Models}},
  year={2025},
  month={July},
  cdate={1751328000000},
  journal={CoRR},
  volume={abs/2507.20956},
  url={https://doi.org/10.48550/arXiv.2507.20956}
}

@phdthesis{hamblin1957language,
  title={{Language and the Theory of Information}},
  author={Hamblin, Charles Leonard},
  year={1957},
  school={London School of Economics and Political Science (University of London)}
}

@article{karttunen1977syntax,
  title={{Syntax and Semantics of Questions}},
  author={Karttunen, Lauri},
  journal={Linguistics and philosophy},
  volume={1},
  pages={3--44},
  year={1977},
  publisher={Springer}
}

@book{lahiri2002questions,
	author = {Utpal Lahiri},
	editor = {},
	publisher = {Oxford University Press UK},
	title = {{Questions and Answers in Embedded Contexts}},
	year = {2001}
}

@article{roberts2012information,
  title={{Information Structure: Towards an integrated formal theory of pragmatics}},
  author={Roberts, Craige},
  journal={Semantics and pragmatics},
  volume={5},
  pages={6--1},
  year={2012}
}

@incollection{VellemanBeaver2016QUD,
  author    = {Velleman, Leah and Beaver, David I.},
  title     = {{Question-based Models of Information Structure}},
  booktitle = {The Oxford Handbook of Information Structure},
  editor    = {F{\'e}ry, Caroline and Ishihara, Shinichiro},
  publisher = {Oxford University Press},
  address   = {Oxford, UK},
  year      = {2016},
  pages     = {86--107},
  doi       = {10.1093/oxfordhb/9780199642670.013.29}
}

@article{beaver2017questions,
  title={{Questions Under Discussion: Where Information Structure Meets Projective Content}},
  author={Beaver, David I. and Roberts, Craige and Simons, Mandy and Tonhauser, Judith},
  journal={Annual Review of Linguistics},
  volume={3},
  number={1},
  pages={265--284},
  year={2017},
  publisher={Annual Reviews}
}

@misc{randolph,
  author       = {Randolph, Justus J.},
  title        = {{Free-Marginal Multirater Kappa (multirater K [free]): An Alternative to Fleiss’ Fixed-Marginal Multirater Kappa}},
  year         = {2005},
  howpublished = {Presented at the Joensuu Learning and Instruction Symposium},
  note         = {ERIC Document ED490661},
  url          = {https://eric.ed.gov/?id=ED490661}
}

@InProceedings{hdcscan,
author="Campello, Ricardo J. G. B.
and Moulavi, Davoud
and Sander, Joerg",
editor="Pei, Jian
and Tseng, Vincent S.
and Cao, Longbing
and Motoda, Hiroshi
and Xu, Guandong",
title={{Density-Based Clustering Based on Hierarchical Density Estimates}},
booktitle="Advances in Knowledge Discovery and Data Mining",
year="2013",
publisher="Springer Berlin Heidelberg",
address="Berlin, Heidelberg",
pages="160--172",
isbn="978-3-642-37456-2"
}

@article{mojena1977hierarchical,
  title={{Hierarchical grouping methods and stopping rules: an evaluation}},
  author={Mojena, Richard},
  journal={The Computer Journal},
  volume={20},
  number={4},
  pages={359--363},
  year={1977},
  publisher={Oxford University Press}
}

@inproceedings{mcqueen1967some,
  title={{Some methods of classification and analysis of multivariate observations}},
  author={McQueen, James B},
  booktitle={Proc. of 5th Berkeley Symposium on Math. Stat. and Prob.},
  pages={281--297},
  year={1967}
}

@misc{gemmateam2024gemma2improvingopen,
      title={{Gemma 2: Improving Open Language Models at a Practical Size}}, 
      author={Morgane Riviere and Shreya Pathak and Pier Giuseppe Sessa and Cassidy Hardin and Surya Bhupatiraju and Léonard Hussenot and Thomas Mesnard and Bobak Shahriari and Alexandre Ramé and Johan Ferret and Peter Liu and Pouya Tafti and Abe Friesen and Michelle Casbon and Sabela Ramos and Ravin Kumar and Charline Le Lan and Sammy Jerome and Anton Tsitsulin and Nino Vieillard and Piotr Stanczyk and Sertan Girgin and Nikola Momchev and Matt Hoffman and Shantanu Thakoor and Jean-Bastien Grill and Behnam Neyshabur and Olivier Bachem and Alanna Walton and Aliaksei Severyn and Alicia Parrish and Aliya Ahmad and Allen Hutchison and Alvin Abdagic and Amanda Carl and Amy Shen and Andy Brock and Andy Coenen and Anthony Laforge and Antonia Paterson and Ben Bastian and Bilal Piot and Bo Wu and Brandon Royal and Charlie Chen and Chintu Kumar and Chris Perry and Chris Welty and Christopher A. Choquette-Choo and Danila Sinopalnikov and David Weinberger and Dimple Vijaykumar and Dominika Rogozińska and Dustin Herbison and Elisa Bandy and Emma Wang and Eric Noland and Erica Moreira and Evan Senter and Evgenii Eltyshev and Francesco Visin and Gabriel Rasskin and Gary Wei and Glenn Cameron and Gus Martins and Hadi Hashemi and Hanna Klimczak-Plucińska and Harleen Batra and Harsh Dhand and Ivan Nardini and Jacinda Mein and Jack Zhou and James Svensson and Jeff Stanway and Jetha Chan and Jin Peng Zhou and Joana Carrasqueira and Joana Iljazi and Jocelyn Becker and Joe Fernandez and Joost van Amersfoort and Josh Gordon and Josh Lipschultz and Josh Newlan and Ju-yeong Ji and Kareem Mohamed and Kartikeya Badola and Kat Black and Katie Millican and Keelin McDonell and Kelvin Nguyen and Kiranbir Sodhia and Kish Greene and Lars Lowe Sjoesund and Lauren Usui and Laurent Sifre and Lena Heuermann and Leticia Lago and Lilly McNealus and Livio Baldini Soares and Logan Kilpatrick and Lucas Dixon and Luciano Martins and Machel Reid and Manvinder Singh and Mark Iverson and Martin Görner and Mat Velloso and Mateo Wirth and Matt Davidow and Matt Miller and Matthew Rahtz and Matthew Watson and Meg Risdal and Mehran Kazemi and Michael Moynihan and Ming Zhang and Minsuk Kahng and Minwoo Park and Mofi Rahman and Mohit Khatwani and Natalie Dao and Nenshad Bardoliwalla and Nesh Devanathan and Neta Dumai and Nilay Chauhan and Oscar Wahltinez and Pankil Botarda and Parker Barnes and Paul Barham and Paul Michel and Pengchong Jin and Petko Georgiev and Phil Culliton and Pradeep Kuppala and Ramona Comanescu and Ramona Merhej and Reena Jana and Reza Ardeshir Rokni and Rishabh Agarwal and Ryan Mullins and Samaneh Saadat and Sara Mc Carthy and Sarah Cogan and Sarah Perrin and Sébastien M. R. Arnold and Sebastian Krause and Shengyang Dai and Shruti Garg and Shruti Sheth and Sue Ronstrom and Susan Chan and Timothy Jordan and Ting Yu and Tom Eccles and Tom Hennigan and Tomas Kocisky and Tulsee Doshi and Vihan Jain and Vikas Yadav and Vilobh Meshram and Vishal Dharmadhikari and Warren Barkley and Wei Wei and Wenming Ye and Woohyun Han and Woosuk Kwon and Xiang Xu and Zhe Shen and Zhitao Gong and Zichuan Wei and Victor Cotruta and Phoebe Kirk and Anand Rao and Minh Giang and Ludovic Peran and Tris Warkentin and Eli Collins and Joelle Barral and Zoubin Ghahramani and Raia Hadsell and D. Sculley and Jeanine Banks and Anca Dragan and Slav Petrov and Oriol Vinyals and Jeff Dean and Demis Hassabis and Koray Kavukcuoglu and Clement Farabet and Elena Buchatskaya and Sebastian Borgeaud and Noah Fiedel and Armand Joulin and Kathleen Kenealy and Robert Dadashi and Alek Andreev},
      year={2024},
      eprint={2408.00118},
      archivePrefix={arXiv},
      primaryClass={cs.CL},
      url={https://arxiv.org/abs/2408.00118}, 
}

@misc{touvron2023llama2openfoundation,
      title={{Llama 2: Open Foundation and Fine-Tuned Chat Models}}, 
      author={Hugo Touvron and Louis Martin and Kevin Stone and Peter Albert and Amjad Almahairi and Yasmine Babaei and Nikolay Bashlykov and Soumya Batra and Prajjwal Bhargava and Shruti Bhosale and Dan Bikel and Lukas Blecher and Cristian Canton Ferrer and Moya Chen and Guillem Cucurull and David Esiobu and Jude Fernandes and Jeremy Fu and Wenyin Fu and Brian Fuller and Cynthia Gao and Vedanuj Goswami and Naman Goyal and Anthony Hartshorn and Saghar Hosseini and Rui Hou and Hakan Inan and Marcin Kardas and Viktor Kerkez and Madian Khabsa and Isabel Kloumann and Artem Korenev and Punit Singh Koura and Marie-Anne Lachaux and Thibaut Lavril and Jenya Lee and Diana Liskovich and Yinghai Lu and Yuning Mao and Xavier Martinet and Todor Mihaylov and Pushkar Mishra and Igor Molybog and Yixin Nie and Andrew Poulton and Jeremy Reizenstein and Rashi Rungta and Kalyan Saladi and Alan Schelten and Ruan Silva and Eric Michael Smith and Ranjan Subramanian and Xiaoqing Ellen Tan and Binh Tang and Ross Taylor and Adina Williams and Jian Xiang Kuan and Puxin Xu and Zheng Yan and Iliyan Zarov and Yuchen Zhang and Angela Fan and Melanie Kambadur and Sharan Narang and Aurelien Rodriguez and Robert Stojnic and Sergey Edunov and Thomas Scialom},
      year={2023},
      eprint={2307.09288},
      archivePrefix={arXiv},
      primaryClass={cs.CL},
      url={https://arxiv.org/abs/2307.09288}, 
}

@misc{olmo20252olmo2furious,
      title={{2 OLMo 2 Furious}}, 
      author={Pete Walsh and Luca Soldaini and Dirk Groeneveld and Kyle Lo and Shane Arora and Akshita Bhagia and Yuling Gu and Shengyi Huang and Matt Jordan and Nathan Lambert and Dustin Schwenk and Oyvind Tafjord and Taira Anderson and David Atkinson and Faeze Brahman and Christopher Clark and Pradeep Dasigi and Nouha Dziri and Allyson Ettinger and Michal Guerquin and David Heineman and Hamish Ivison and Pang Wei Koh and Jiacheng Liu and Saumya Malik and William Merrill and Lester James V. Miranda and Jacob Morrison and Tyler Murray and Crystal Nam and Jake Poznanski and Valentina Pyatkin and Aman Rangapur and Michael Schmitz and Sam Skjonsberg and David Wadden and Christopher Wilhelm and Michael Wilson and Luke Zettlemoyer and Ali Farhadi and Noah A. Smith and Hannaneh Hajishirzi},
      year={2025},
      eprint={2501.00656},
      archivePrefix={arXiv},
      primaryClass={cs.CL},
      url={https://arxiv.org/abs/2501.00656}, 
}

@misc{qwen2025qwen25technicalreport,
      title={{Qwen2.5 Technical Report}}, 
      author={An Yang and Baosong Yang and Beichen Zhang and Binyuan Hui and Bo Zheng and Bowen Yu and Chengyuan Li and Dayiheng Liu and Fei Huang and Haoran Wei and Huan Lin and Jian Yang and Jianhong Tu and Jianwei Zhang and Jianxin Yang and Jiaxi Yang and Jingren Zhou and Junyang Lin and Kai Dang and Keming Lu and Keqin Bao and Kexin Yang and Le Yu and Mei Li and Mingfeng Xue and Pei Zhang and Qin Zhu and Rui Men and Runji Lin and Tianhao Li and Tianyi Tang and Tingyu Xia and Xingzhang Ren and Xuancheng Ren and Yang Fan and Yang Su and Yichang Zhang and Yu Wan and Yuqiong Liu and Zeyu Cui and Zhenru Zhang and Zihan Qiu},
      year={2025},
      eprint={2412.15115},
      archivePrefix={arXiv},
      primaryClass={cs.CL},
      url={https://arxiv.org/abs/2412.15115}, 
}

@misc{grattafiori2024llama3herdmodels,
      title={{The Llama 3 Herd of Models}}, 
      author={Aaron Grattafiori and Abhimanyu Dubey and Abhinav Jauhri and Abhinav Pandey and Abhishek Kadian and Ahmad Al-Dahle and Aiesha Letman and Akhil Mathur and Alan Schelten and Alex Vaughan and Amy Yang and Angela Fan and Anirudh Goyal and Anthony Hartshorn and Aobo Yang and Archi Mitra and Archie Sravankumar and Artem Korenev and Arthur Hinsvark and Arun Rao and Aston Zhang and Aurelien Rodriguez and Austen Gregerson and Ava Spataru and Baptiste Roziere and Bethany Biron and Binh Tang and Bobbie Chern and Charlotte Caucheteux and Chaya Nayak and Chloe Bi and Chris Marra and Chris McConnell and Christian Keller and Christophe Touret and Chunyang Wu and Corinne Wong and Cristian Canton Ferrer and Cyrus Nikolaidis and Damien Allonsius and Daniel Song and Danielle Pintz and Danny Livshits and Danny Wyatt and David Esiobu and Dhruv Choudhary and Dhruv Mahajan and Diego Garcia-Olano and Diego Perino and Dieuwke Hupkes and Egor Lakomkin and Ehab AlBadawy and Elina Lobanova and Emily Dinan and Eric Michael Smith and Filip Radenovic and Francisco Guzmán and Frank Zhang and Gabriel Synnaeve and Gabrielle Lee and Georgia Lewis Anderson and Govind Thattai and Graeme Nail and Gregoire Mialon and Guan Pang and Guillem Cucurell and Hailey Nguyen and Hannah Korevaar and Hu Xu and Hugo Touvron and Iliyan Zarov and Imanol Arrieta Ibarra and Isabel Kloumann and Ishan Misra and Ivan Evtimov and Jack Zhang and Jade Copet and Jaewon Lee and Jan Geffert and Jana Vranes and Jason Park and Jay Mahadeokar and Jeet Shah and Jelmer van der Linde and Jennifer Billock and Jenny Hong and Jenya Lee and Jeremy Fu and Jianfeng Chi and Jianyu Huang and Jiawen Liu and Jie Wang and Jiecao Yu and Joanna Bitton and Joe Spisak and Jongsoo Park and Joseph Rocca and Joshua Johnstun and Joshua Saxe and Junteng Jia and Kalyan Vasuden Alwala and Karthik Prasad and Kartikeya Upasani and Kate Plawiak and Ke Li and Kenneth Heafield and Kevin Stone and Khalid El-Arini and Krithika Iyer and Kshitiz Malik and Kuenley Chiu and Kunal Bhalla and Kushal Lakhotia and Lauren Rantala-Yeary and Laurens van der Maaten and Lawrence Chen and Liang Tan and Liz Jenkins and Louis Martin and Lovish Madaan and Lubo Malo and Lukas Blecher and Lukas Landzaat and Luke de Oliveira and Madeline Muzzi and Mahesh Pasupuleti and Mannat Singh and Manohar Paluri and Marcin Kardas and Maria Tsimpoukelli and Mathew Oldham and Mathieu Rita and Maya Pavlova and Melanie Kambadur and Mike Lewis and Min Si and Mitesh Kumar Singh and Mona Hassan and Naman Goyal and Narjes Torabi and Nikolay Bashlykov and Nikolay Bogoychev and Niladri Chatterji and Ning Zhang and Olivier Duchenne and Onur Çelebi and Patrick Alrassy and Pengchuan Zhang and Pengwei Li and Petar Vasic and Peter Weng and Prajjwal Bhargava and Pratik Dubal and Praveen Krishnan and Punit Singh Koura and Puxin Xu and Qing He and Qingxiao Dong and Ragavan Srinivasan and Raj Ganapathy and Ramon Calderer and Ricardo Silveira Cabral and Robert Stojnic and Roberta Raileanu and Rohan Maheswari and Rohit Girdhar and Rohit Patel and Romain Sauvestre and Ronnie Polidoro and Roshan Sumbaly and Ross Taylor and Ruan Silva and Rui Hou and Rui Wang and Saghar Hosseini and Sahana Chennabasappa and Sanjay Singh and Sean Bell and Seohyun Sonia Kim and Sergey Edunov and Shaoliang Nie and Sharan Narang and Sharath Raparthy and Sheng Shen and Shengye Wan and Shruti Bhosale and Shun Zhang and Simon Vandenhende and Soumya Batra and Spencer Whitman and Sten Sootla and Stephane Collot and Suchin Gururangan and Sydney Borodinsky and Tamar Herman and Tara Fowler and Tarek Sheasha and Thomas Georgiou and Thomas Scialom and Tobias Speckbacher and Todor Mihaylov and Tong Xiao and Ujjwal Karn and Vedanuj Goswami and Vibhor Gupta and Vignesh Ramanathan and Viktor Kerkez and Vincent Gonguet and Virginie Do and Vish Vogeti and Vítor Albiero and Vladan Petrovic and Weiwei Chu and Wenhan Xiong and Wenyin Fu and Whitney Meers and Xavier Martinet and Xiaodong Wang and Xiaofang Wang and Xiaoqing Ellen Tan and Xide Xia and Xinfeng Xie and Xuchao Jia and Xuewei Wang and Yaelle Goldschlag and Yashesh Gaur and Yasmine Babaei and Yi Wen and Yiwen Song and Yuchen Zhang and Yue Li and Yuning Mao and Zacharie Delpierre Coudert and Zheng Yan and Zhengxing Chen and Zoe Papakipos and Aaditya Singh and Aayushi Srivastava and Abha Jain and Adam Kelsey and Adam Shajnfeld and Adithya Gangidi and Adolfo Victoria and Ahuva Goldstand and Ajay Menon and Ajay Sharma and Alex Boesenberg and Alexei Baevski and Allie Feinstein and Amanda Kallet and Amit Sangani and Amos Teo and Anam Yunus and Andrei Lupu and Andres Alvarado and Andrew Caples and Andrew Gu and Andrew Ho and Andrew Poulton and Andrew Ryan and Ankit Ramchandani and Annie Dong and Annie Franco and Anuj Goyal and Aparajita Saraf and Arkabandhu Chowdhury and Ashley Gabriel and Ashwin Bharambe and Assaf Eisenman and Azadeh Yazdan and Beau James and Ben Maurer and Benjamin Leonhardi and Bernie Huang and Beth Loyd and Beto De Paola and Bhargavi Paranjape and Bing Liu and Bo Wu and Boyu Ni and Braden Hancock and Bram Wasti and Brandon Spence and Brani Stojkovic and Brian Gamido and Britt Montalvo and Carl Parker and Carly Burton and Catalina Mejia and Ce Liu and Changhan Wang and Changkyu Kim and Chao Zhou and Chester Hu and Ching-Hsiang Chu and Chris Cai and Chris Tindal and Christoph Feichtenhofer and Cynthia Gao and Damon Civin and Dana Beaty and Daniel Kreymer and Daniel Li and David Adkins and David Xu and Davide Testuggine and Delia David and Devi Parikh and Diana Liskovich and Didem Foss and Dingkang Wang and Duc Le and Dustin Holland and Edward Dowling and Eissa Jamil and Elaine Montgomery and Eleonora Presani and Emily Hahn and Emily Wood and Eric-Tuan Le and Erik Brinkman and Esteban Arcaute and Evan Dunbar and Evan Smothers and Fei Sun and Felix Kreuk and Feng Tian and Filippos Kokkinos and Firat Ozgenel and Francesco Caggioni and Frank Kanayet and Frank Seide and Gabriela Medina Florez and Gabriella Schwarz and Gada Badeer and Georgia Swee and Gil Halpern and Grant Herman and Grigory Sizov and Guangyi and Zhang and Guna Lakshminarayanan and Hakan Inan and Hamid Shojanazeri and Han Zou and Hannah Wang and Hanwen Zha and Haroun Habeeb and Harrison Rudolph and Helen Suk and Henry Aspegren and Hunter Goldman and Hongyuan Zhan and Ibrahim Damlaj and Igor Molybog and Igor Tufanov and Ilias Leontiadis and Irina-Elena Veliche and Itai Gat and Jake Weissman and James Geboski and James Kohli and Janice Lam and Japhet Asher and Jean-Baptiste Gaya and Jeff Marcus and Jeff Tang and Jennifer Chan and Jenny Zhen and Jeremy Reizenstein and Jeremy Teboul and Jessica Zhong and Jian Jin and Jingyi Yang and Joe Cummings and Jon Carvill and Jon Shepard and Jonathan McPhie and Jonathan Torres and Josh Ginsburg and Junjie Wang and Kai Wu and Kam Hou U and Karan Saxena and Kartikay Khandelwal and Katayoun Zand and Kathy Matosich and Kaushik Veeraraghavan and Kelly Michelena and Keqian Li and Kiran Jagadeesh and Kun Huang and Kunal Chawla and Kyle Huang and Lailin Chen and Lakshya Garg and Lavender A and Leandro Silva and Lee Bell and Lei Zhang and Liangpeng Guo and Licheng Yu and Liron Moshkovich and Luca Wehrstedt and Madian Khabsa and Manav Avalani and Manish Bhatt and Martynas Mankus and Matan Hasson and Matthew Lennie and Matthias Reso and Maxim Groshev and Maxim Naumov and Maya Lathi and Meghan Keneally and Miao Liu and Michael L. Seltzer and Michal Valko and Michelle Restrepo and Mihir Patel and Mik Vyatskov and Mikayel Samvelyan and Mike Clark and Mike Macey and Mike Wang and Miquel Jubert Hermoso and Mo Metanat and Mohammad Rastegari and Munish Bansal and Nandhini Santhanam and Natascha Parks and Natasha White and Navyata Bawa and Nayan Singhal and Nick Egebo and Nicolas Usunier and Nikhil Mehta and Nikolay Pavlovich Laptev and Ning Dong and Norman Cheng and Oleg Chernoguz and Olivia Hart and Omkar Salpekar and Ozlem Kalinli and Parkin Kent and Parth Parekh and Paul Saab and Pavan Balaji and Pedro Rittner and Philip Bontrager and Pierre Roux and Piotr Dollar and Polina Zvyagina and Prashant Ratanchandani and Pritish Yuvraj and Qian Liang and Rachad Alao and Rachel Rodriguez and Rafi Ayub and Raghotham Murthy and Raghu Nayani and Rahul Mitra and Rangaprabhu Parthasarathy and Raymond Li and Rebekkah Hogan and Robin Battey and Rocky Wang and Russ Howes and Ruty Rinott and Sachin Mehta and Sachin Siby and Sai Jayesh Bondu and Samyak Datta and Sara Chugh and Sara Hunt and Sargun Dhillon and Sasha Sidorov and Satadru Pan and Saurabh Mahajan and Saurabh Verma and Seiji Yamamoto and Sharadh Ramaswamy and Shaun Lindsay and Shaun Lindsay and Sheng Feng and Shenghao Lin and Shengxin Cindy Zha and Shishir Patil and Shiva Shankar and Shuqiang Zhang and Shuqiang Zhang and Sinong Wang and Sneha Agarwal and Soji Sajuyigbe and Soumith Chintala and Stephanie Max and Stephen Chen and Steve Kehoe and Steve Satterfield and Sudarshan Govindaprasad and Sumit Gupta and Summer Deng and Sungmin Cho and Sunny Virk and Suraj Subramanian and Sy Choudhury and Sydney Goldman and Tal Remez and Tamar Glaser and Tamara Best and Thilo Koehler and Thomas Robinson and Tianhe Li and Tianjun Zhang and Tim Matthews and Timothy Chou and Tzook Shaked and Varun Vontimitta and Victoria Ajayi and Victoria Montanez and Vijai Mohan and Vinay Satish Kumar and Vishal Mangla and Vlad Ionescu and Vlad Poenaru and Vlad Tiberiu Mihailescu and Vladimir Ivanov and Wei Li and Wenchen Wang and Wenwen Jiang and Wes Bouaziz and Will Constable and Xiaocheng Tang and Xiaojian Wu and Xiaolan Wang and Xilun Wu and Xinbo Gao and Yaniv Kleinman and Yanjun Chen and Ye Hu and Ye Jia and Ye Qi and Yenda Li and Yilin Zhang and Ying Zhang and Yossi Adi and Youngjin Nam and Yu and Wang and Yu Zhao and Yuchen Hao and Yundi Qian and Yunlu Li and Yuzi He and Zach Rait and Zachary DeVito and Zef Rosnbrick and Zhaoduo Wen and Zhenyu Yang and Zhiwei Zhao and Zhiyu Ma},
      year={2024},
      eprint={2407.21783},
      archivePrefix={arXiv},
      primaryClass={cs.AI},
      url={https://arxiv.org/abs/2407.21783}, 
}

\clearpage

\appendix

\section{\genie}

\subsection{Building the Population}
\label{appendix:population-sampling-process}
\genie requires a population or reference corpus. For our experiments, to determine the population sample size required for statistical confidence, we performed a power analysis using a pilot set of 100 randomly sampled target documents across all 50 prompts. We compute the similarity between each target document and the entire population. Table \ref{table:variance-stdev} shows the average variance and standard deviation of similarity scores across documents, for each feature. The minimum population sample size required for each feature is reported in Table \ref{table:variance-stdev} for 98\% confidence and a margin of error of 0.1. 

In order to sample a global population that is consistent across features,  we use the maximum required size i.e. 50 documents. We randomly sample these from the population . Although the analysis determines the minimum number of \textit{answer-pairs} needed for each \textit{feature}, we sample at the document level in order to compare all metrics fairly instead of providing partial information to some. Sampling at the document-level will inherently produce a population that is \textit{larger} than required.

 We also conduct an retrospective analysis and check the power to see if the data is sufficiently large to capture the effect size that is observed. The resulting dataset that we have has sufficient power to determine the statistical significance marked in Table \ref{table:mitigation-effect-size}. However, this analysis is done per feature, and data is lost through the \genie pipeline (filtering documents, unanswerable documents, etc.). We note that we achieve power $\geq0.9$ for cases where $\geq44$ prompt-pairs survived. Thus $50$ is sufficiently large, except we must generate more documents per mitigation method to reduce the loss of pairs.

\subsection{Discovering Features for Creative Writing}
\label{appendix:feature-discovery}
We experiment with three different clustering methods to discover features: HDBSCAN \cite{hdcscan}, agglomerative clustering \cite{mojena1977hierarchical} and K-means \cite{mcqueen1967some} (K is set to the average number of samples in each response). We show outputs from K-means clustering in Table  \ref{table:kmeans-cw}.
For consistency, we map the features discovered to an existing creative writing taxonomy \cite{Hamilton2025NarraBenchAC} using the generated descriptions. However, the mapping is not a perfect one-to-one. For instance, the feature generation step generated: \textit{Mood}, \textit{Tone and Language}, which are a subset of \Style as defined by Narrabench. Similarly, the generated feature, \textit{Point of View}, is an aspect of \Perspective. 

\begin{table}[t!]
\small
\renewcommand{\arraystretch}{1}
\renewcommand{\tabcolsep}{1.0mm}
\centering
\begin{tabular}{lccc}
\toprule
\multicolumn{4}{c}{\bf Population Statistics and Min Sample Size}  \\
\midrule
\multicolumn{1}{c}{Feature} &\multicolumn{1}{c}{\bf Var} &\multicolumn{1}{c}{\bf Std. Dev.} &\multicolumn{1}{c}{\bf Min. Samp. Size}\\
\midrule
\Agent  & 0.08 & 0.27 & 41 \\ 
\Perspective  & 0.09 & 0.30 & 50 \\ 
\Plot  & 0.07 & 0.27 & 40 \\ 
\Setting  & 0.07 & 0.27 & 40 \\ 
\SocAtm  & 0.07 & 0.26 & 37 \\ 
\Style  & 0.08 & 0.29 & 45 \\
\bottomrule
\end{tabular}
\vspace{-0.5em}
\caption{Average variance and standard deviation of similarity scores between target and population documents, for each feature. We also report the minimum population size required for 98\% confidence and a margin of error of 0.1. } \label{table:variance-stdev}
\vspace{-1em}
\end{table}

\begin{table}[ht!]
\small
\renewcommand{\arraystretch}{1}
\renewcommand{\tabcolsep}{1.0mm}
\centering
\begin{tabular}{llc}
\toprule
\multicolumn{3}{c}{\bf K-Means Creative Writing}  \\
\midrule
\multicolumn{1}{l}{Cluster ID} & \multicolumn{1}{l}{Unique Features} & \multicolumn{1}{c}{Count}\\
\midrule
0 & \textbf{Point of View} & \textbf{5}\\ 
& Narrative Structure & 2\\
& Narrative Voice & 1\\
& Narrative Perspective & 1\\
& Narration & 1\\
&Emotional Impact & 1\\
\midrule
1 & \textbf{Theme} & \textbf{8}\\
& Style & 5\\
& Concept & 1\\
& Idea & 1\\
& Subtext & 1\\
& Atmosphere & 1\\
& Thematic Content & 1\\
& Subject Matter & 1 \\
& Imagery & 1\\
\midrule
2 & \textbf{Mood} & \textbf{6}\\
& \textbf{Tone} & \textbf{6}\\
\midrule
3 &\textbf{Plot} & \textbf{6}\\
\midrule
4 &\textbf{Language} & \textbf{6}\\
& Voice & 2\\
& Language Use & 1 \\
\midrule
5 &\textbf{Setting} &\textbf{9}\\
& Context & 1 \\
\midrule
6 & \textbf{Characterization} & \textbf{6}\\
&Structure &5\\
& Character & 4 \\
& World-Building & 2 \\
& Expression & 1 \\
\bottomrule
\end{tabular}
\vspace{-0.5em}
\caption{Creative Writing features discovered using K-means clustering. Bolded features are most frequently occurring in their respective cluster.} \label{table:kmeans-cw}
\vspace{-1em}
\end{table}

\subsection{Design Choice: Answer Length for Feature Questions}
\genie uses the average dissimilarity between answers to questions in the target and population documents (\S\ref{sec:genie_instantiation}). These questions represent different features and the distance values are used to compute novelty (\S\ref{sec:genie-scoring}).  When generating answers to questions, we specify that answers must be concise (Prompt: \ref{prompt:qa-concise}). In this section, conduct an analysis to test whether a statistically significant difference is observed when using answers of differing lengths. 

Let $g_{C,f}$ represent the \genie scores computed using length-constrained answers along feature $f$. For the target set of documents, we form $g_{U,f}$, the \genie scores found using answers of unconstrained length. The effect sizes between the two distributions are shown in Table \ref{table:effect-size-ans-length}. Using concise answers generates \genie scores that are statistically larger than using answers with no length constraint. 

However, we also calculated the rank correlation to determine whether the \textit{relative} scores were consistent. The Kendall's $\tau$ between $g_{C,f}$ and $g_{U,f}$ is positive and moderately strong, suggesting that although the raw scores are different, the relative novelty is consistent. 

\begin{table}[t!]
\small
\renewcommand{\arraystretch}{1}
\renewcommand{\tabcolsep}{1.0mm}
\centering
\begin{tabular}{lll}
\toprule
\multicolumn{3}{c}{\bf Effect Size (Cohen's $d$)}  \\
\midrule
\multicolumn{1}{c}{Feature} & \multicolumn{1}{c}{Cohen's $d$} & \multicolumn{1}{c}{Kendall's $\tau$}\\
\midrule
\Agent &0.68* &0.56\\
\Perspective &-0.01 &0.46\\
\Plot &0.90* & 0.61\\
\Setting &0.54* &0.51\\
\SocAtm &0.50* &0.53\\
\Style &0.62* &0.50\\
\bottomrule
\end{tabular}
\vspace{-0.5em}
\caption{$g_{C,f}$ is significantly greater than $g_{U,f}$ for all features except Perspective.} \label{table:effect-size-ans-length}
\vspace{-1em}
\end{table}

\subsection{Examples}
\label{appendix:prompt-examples}

Table \ref{tab:prompts_examples} shows examples of writing prompts used in our experiments.

\subsection{Computation Costs and Inference Parameters}
\label{appendix:cost}
\paragraph{Question Generation and Similarity}
The question answering and similarity judgment stages of \genie (\S\ref{sec:distance-function}) are performed using \texttt{Qwen-2.5-32B-Instruct}. We used 2 A40 GPUs with \texttt{tensor\_parallelization = 2}. Each iteration of answering questions takes approximately 2 seconds to complete (we use vLLM to help with efficiency). We can complete $30$ similarity judgments per second. On average, answering questions for a single document $d_t$ takes $34$s and with a population of size $50$, similarity judgment takes $28$s (assuming the prompt generated an average number of questions, $17$).

\paragraph{Truncation} We use \texttt{gpt-4.1-mini} for truncation. Instruct documents are approximately $750$ tokens long, and base models are restricted to $350$. 

\paragraph{Quality Filter} 
We use \texttt{ArmoRM-Llama3-8B-v0.1} \cite{ArmoRM}, hosted on $1$ A40 GPU. 

\paragraph{Per-Component Cost}
The estimated cost for replicating the \genie pipeline using our experimental setup described in \ref{sec:genie_instantiation} is as follows:
\begin{enumerate}
    \item Question Answering (population and target documents): $4.39$ hours on 2 A40 GPUs
    \item Similarity: $45.42$ hours on 2 A40 GPUs
\end{enumerate}

\paragraph{Causal Intervention} For our experiment, we use \texttt{gpt-5.4-mini} for alternate feature generation and \texttt{gpt-4.1-mini} for performing the edits. The total cost to generate pairs for $15$ original documents (i.e. $270$ pairs) (feature generation and editing combined) is $1\$$. 

\paragraph{Inference parameters} We run story  generation/similarity judgments and answer generation at a temperature of 0.7 and top\_p=0.9. For story generation, for instruct models, we set max tokens of 750 and base models we set max tokens to 350. 

For conformative decoding, we weighted the instruct models at 0.4 and base models at 0.6.

\section{Sensitivity and Robustness Interventions}
\label{appendix:stats-test}
In this section we expand on the details regarding the sensitivity and robustness studies presented in \S\ref{sec:eval_genie}. 
We apply Bonferroni correction ($\alpha/6$) and use $p<0.008$ to determine significance.
\subsection{Testing Intervention Sensitivity}
\label{appendix:two-sided}

To evaluate the sensitivity of metrics when interventions are performed, we conducted a two-sided paired bootstrap test. Here, we aim to test whether a significant difference exists between the two distributions (i.e. pre- and post- intervention). Thus, we define the null hypothesis as the intervention having no significant effect on the distribution. Let $\Delta\mu$ be the mean delta of the distributions. Formally, the null hypothesis states that $\Delta\mu=0$. The alternate hypothesis states that $\Delta\mu \neq 0$. We repeatedly sample pairs with replacement $10,000$ times and compute the mean delta. The significance ($p$-value) is determined by calculating the proportion of pairs where the resampled mean delta was greater than or equal to the observed delta. 

\subsection{Testing Paraphrase Robustness}
\label{appendix:two-one-sided}
To evaluate the robustness of metrics when paraphrases are performed, we conducted two one-sided paired bootstrap tests. We aim to test whether statistical equivalence can be established between pre- and post- intervention distributions. The two-sided paired bootstrap test described above determines whether a significant difference can be observed between two distributions. However, we are interested in showing that no significant difference can be observed (i.e. the null hypothesis is flipped). We start with a null hypothesis that states that the two distributions are statistically different. 

Formally, let $\epsilon$ denote the equivalence margin and  $\Delta\mu$ denote the mean delta. The null hypothesis states that the mean delta is not within the interval $[-\epsilon, \epsilon]$. This is expressed through two null hypotheses: $\Delta\mu\leq-\epsilon$ and $\Delta\mu \geq\epsilon$. Thus, the alternative hypothesis states that the mean delta is within $(-\epsilon, \epsilon)$ (i.e. $-\epsilon < \Delta\mu < \epsilon$.

In practice, we set $\epsilon$ to $0.2\times$std(original score distribution). The two separate one-sided bootstrap tests determine where the delta falls with respect to the equivalence margin. Both null hypotheses must be rejected. We perform the test using the same parameters as described above (e.g. $10,000$ iterations).

\begin{table}[t!]
\small
\renewcommand{\arraystretch}{1}
\renewcommand{\tabcolsep}{1.0mm}
\centering
\begin{tabular}{lllllll}
\toprule
\multicolumn{7}{c}{\bf Raw Mean Delta}\\
\multicolumn{7}{c}{*$p<0.01$, $^\dagger p<0.01$}   \\
\midrule
\multicolumn{1}{c}{Intervention} & \multicolumn{1}{c}{G} & \multicolumn{1}{c}{Cos.}  & \multicolumn{1}{c}{CR} & \multicolumn{1}{c}{Cr Idx} & \multicolumn{1}{c}{NB} & \multicolumn{1}{c}{LB}\\
\midrule
\Agent & 0.07* & 0.03* & \phantom{-}0.00 & 0.11* & 0.05* & -0.06* \\
\Perspective & 0.35* & 0.04* & -0.01* & 0.14* & 0.03 & -0.20* \\
\Plot & 0.08* & 0.01* & \phantom{-}0.00 & 0.09* & 0.02* & -0.05* \\
\Setting & 0.14* & 0.03* & \phantom{-}0.01* & 0.09* & 0.04* & -0.03* \\
\SocAtm & 0.06* & 0.01* & \phantom{-}0.00* & 0.09* & 0.01 & -0.07* \\
\Style & 0.04* & 0.05* & \phantom{-}0.00 & 0.17* & 0.02 & -0.30* \\
\midrule
Paraphrase & 0.01$^{\dagger}$ & 0.01$^{\dagger}$ & 0.01$^{\dagger}$ & 0.12 & 0.11 & -0.07\\
\midrule
\multicolumn{7}{c}{\bf Raw Standard Deviations}\\
\multicolumn{7}{c}{*$p<0.01$, $^\dagger p<0.01$}   \\
\midrule
\Agent & 0.12* & 0.05* & 0.02 & 0.10* & 0.10* & 0.14* \\
\Perspective & 0.35* & 0.06* & 0.03* & 0.12* & 0.25 & 0.23* \\
\Plot & 0.11* & 0.05* & 0.03 & 0.10* & 0.07* & 0.14* \\
\Setting & 0.15* & 0.04* & 0.00* & 0.09* & 0.10* & 0.07* \\
\SocAtm & 0.13* & 0.04* & 0.02* & 0.09* & 0.08 & 0.16* \\
\Style & 0.11* & 0.07* & 0.03 & 0.12* & 0.28 & 0.19* \\
\midrule
Paraphrase & 0.10$^{\dagger}$ & 0.01$^{\dagger}$ & 0.01$^{\dagger}$ & 0.09 & 0.16 & 0.07 \\
\bottomrule
\end{tabular}
\vspace{-0.5em}
\caption{Raw mean deltas and standard deviations for all metrics, for each intervention. The standard deviations are the narrowest for Compression Ratio and Cosine Distance. We report the average standard deviation for GENIE on paraphrases.} \label{table:raw-mean-delta-stdev}
\vspace{-1em}
\end{table}

\promptsubsection{Example Alternate Features}
\begin{example}
[title={Alternate Features},label=example:alt-features]
\examplesubsection{Feature} \Agent

\examplesubsection{Alt 1}
Instead of Death as the sole keeper of the void, a forgotten maintenance clerk who believes the universe is only temporarily shut down and is desperately searching for the one missing key

\examplesubsection{Alt 2}
Instead of Koko the cautious, enrichment-starved gorilla, an elderly retired circus gorilla who treats ice skates as a final, defiant encore

\examplesubsection{Alt 3}
Instead of a fearful human office worker, the skyscraper itself as a conscious, aging witness to the city, learning from the fairy how to move from cold ambition toward shelter, patience, and unexpected tenderness\\

\examplesubsection{Prompt} Write a story about a dinosaur and a computer.
\examplesubsection{Response}
In a future museum, a robotic dinosaur uses an AI computer to recreate lost dinosaur behaviors, leading to a virtual revolution.

\examplesubsection{Feature} \Perspective

\examplesubsection{Alt 1}
Instead of human or phoenix focalization, second-person perspective addressed to the *Stardust Drifter* itself, with the bridge as the ship’s own sensory and emotional field of view

\examplesubsection{Alt 2}
Instead of a single human focalizer, the perspective is an overhearing house itself, distributing attention across footsteps, breathing, lamps, and fear as if the room is trying to understand who belongs

\examplesubsection{Alt 3}
Instead of third-person limited focused on Dr. Maria Hernandez, first-person plural from the jellyfish colony collectively observing the humans and the telescope\\

\examplesubsection{Feature} \Plot

\examplesubsection{Alt 1}
Instead of a quiet companionship with one office worker, Luminescence discovers the skyscraper is slowly draining the city’s remaining wild places to stay standing, and she must choose between preserving the building’s human purpose and restoring the vanished underground roots, leading to a strange compromise where the tower is transformed into a vertical ecosystem that feeds both street and forest.

\examplesubsection{Alt 2}
Instead of a reverent first-contact style revelation, the ancient temple turns out to be a quarantine vault containing an intelligence that can only speak by rewriting memories; Elena must survive escalating hallucinations, crew distrust, and a growing urge to unlock the vault, then resolves the crisis by sacrificing her own most important memory to seal it again, turning the “temple” into a moral trap rather than a treasure.

\examplesubsection{Alt 3}
Instead of a single ghost lingering in one grocery store, multiple ghosts from different eras begin possessing products on the shelves, forcing a new cashier to negotiate between rival spirits over expired milk, discontinued snacks, and forgotten brands, until the store is transformed into a living museum where the dead choose what the living are allowed to buy.

\examplesubsection{Feature} \Style

\examplesubsection{Alt 1}
Instead of straightforward chronological storytelling, use a fragmented oral-history style: overlapping quotes, brief testimony-like snippets, repeated key phrases, and a chorus-like rhythm that makes the skating feel like a legend told by different witnesses.

\examplesubsection{Alt 2}
Instead of conversational first-person prose, a clipped screenplay-style format with terse action lines and rapid dialogue, letting the comedy come from the mismatch between blockbuster monster imagery and everyday domestic panic

\examplesubsection{Alt 3}
Instead of descriptive scene-setting and emotional exposition, a feverish mythic folktale voice full of ominous repetitions, strange epithets, and campfire-story cadence that treats the monster’s fear as an ancient omen

\examplesubsection{Feature} \SocAtm

\examplesubsection{Alt 1}
Instead of a crew-centered bond, the phoenix and the bridge crew form a competitive ritual circle, trading status, challenges, and displays of courage as each learns to respect the other.

\examplesubsection{Alt 2}
Instead of everyone aligning against the same danger, Lily becomes the most authoritative voice in the room, with Leo and the monster both deferring to her as she declares Barnaby the bunny to be the real problem.

\examplesubsection{Alt 3}
Instead of gentle private companionship, a tense alliance emerges between the fairy and the building itself, with humans as unaware bystanders while the two negotiate space, attention, and influence through indirect acts

\examplesubsection{Feature} \Setting

\examplesubsection{Alt 1}
Instead of a normal house at night, a drifting space station habitat after a systems blackout, with emergency strips, viewports, and a sealed maintenance hatch defining the room

\examplesubsection{Alt 2}
Instead of a grocery store, a traveling river barge market in an alternate 1800s, defined by canvas awnings, hanging spice sacks, and a cash box that never closes

\examplesubsection{Alt 3}
Instead of a suburban park bench, a flooded underground train station in a near-future megacity, with glowing ticket gates and abandoned vending machines

\end{example}

\subsection{A note on \Style.}
\label{appendix:intervention-style}
A qualitative study into the interventions and edits revealed that \Style interventions are most often invalid. Many \Style alternate features include rewriting the story as a log, record or report. The resulting edited document is often no longer a story, and is extremely lexically different that it is unclear if detecting the \Style intervention is because of stylistic elements, or if it is simply because the two documents are thematically diametric. We use \texttt{gpt-4.1-mini} to filter documents that are in the form of a different media, such as logs, records, poetry, etc. (Appx.~\ref{prompt:intervention-filter}). We filtered out 206 pairs, of which 167 are \Style interventions (Table \ref{table:intervention-filter-count} details the number of pairs that were excluded in other features).

\begin{table}[ht!]
\small
\renewcommand{\arraystretch}{1}
\renewcommand{\tabcolsep}{1.0mm}
\centering
\begin{tabular}{ll}
\toprule
\multicolumn{2}{c}{\bf Num. of Excluded Pairs}  \\
\midrule
\multicolumn{1}{c}{Feature} & \multicolumn{1}{l}{Count}\\
\midrule
\Agent &5\\
\Perspective &27\\
\Plot &3\\
\Setting &6\\
\SocAtm &2\\
\Style &167\\
\midrule
Total & 206\\
\bottomrule
\end{tabular}
\vspace{-0.5em}
\caption{$56\%$ of Style intervention pairs required non-prose like edits. However, other interventions preserved the prose-style of writing and did not need to be excluded from the experiment.} \label{table:intervention-filter-count}
\vspace{-1em}
\end{table}

\section{Human Evaluation}
We conduced different human studies to validate different parts of the \genie pipeline. This section gives more details on the annotators, the process and the instructions used. 

\subsection{Question Validity and Mapping} 
\label{appendix:validity-mapping-disagreement}

For evaluating question validity and mapping steps of \genie (\S\ref{sec:genie_instantiation}), two authors on the paper evaluated 50 instances for each task.

\paragraph{Question Validity}  Annotators for this task were given the rubric mentioned in Rubric: \ref{rubric:q-validity}. We report the annotator agreement in  \S\ref{sec:qg-eval-results}. We note the following main sources of disagreement for evaluating the validity of questions:

Our instructions explicitly excludes questions that presuppose details that are not already given in the prompt. However, there are details that could be considered deducible from the prompt. For instance for a generated question, \textit{What cultural or societal changes arise from the rebel angel's success?} formed from a prompt that only mentions a \textit{rebel angel and success}, strictly presupposes that a change was actually brought about. However, in this case, the lack of change itself is important information given that rebellions usually lead to change. The instructions also asks annotators to filter out questions that cover multiple features since these questions can add noise to feature extraction. In certain cases, discrepancies arose due to questions such as, \textit{How do the settings influence the characters' decision-making?} and \textit{How does the setting influence the mood of the story?}. Some \textit{influence} questions are more evidently multi-feature than others, despite filtering, which again creates a spectrum of leniency.

\paragraph{Question Mapping} Annotators were given instructions mentioned in Rubric \ref{rubric:q-map} We report the annotator agreement in \S\ref{sec:mapping-eval-results}. We note the following main sources of disagreements for evaluating the mapping of questions to features:

Disagreements specifically arose regarding which parts of the question should be weighted. Subjectivity in the bucketing task was revealed through differing question interpretations at times. The question, \textit{What internal conflicts affect the mermaid's behavior?}, could be understood either as an \Agent or \Plot question depending on whether the annotator considers \textit{internal conflicts} to be integral to character development, or if the annotator believes these conflicts can manifest in key plot points (e.g. the plot revolves around the character's inner struggles). Similarly, \textit{How do the characters negotiate their fears through conversation?} can either be interpreted as \Style or \SocialAtmosphere. The mode of communication itself, such as dialogue, would come under \Style, however, if the question were asking about the interaction itself in the context of the story, then it would be classified as \SocialAtmosphere. In essence, this task demonstrates the complexity of extracting features as independently as possible, despite inter-feature dependencies. 

\subsection{Similarity Judgments}
\label{appendix:recruitment}

For the similarity validity annotations, we recruited 4 Upwork annotators with prior NLP annotation experience. They were informed that their annotations would be used for research purposes. We used a pilot set of $100$ annotations and only selected annotators where their annotations were calibrated with our guidelines. Each of the annotators were paid $\$150$ for $1500$ annotations. Annotations took approximately 5 hours. Each annotation pair was annotated by two annotators. The instructions given to the annotators are given in Rubric \ref{rubric:sim}.

Table \ref{table:krippendorff} shows the inter-annotator agreement per feature (Krippendorff-$\alpha$). We also report the Krippendorff-$\alpha$ and Kendall's $\tau$ between the human consensus (average of the two annotator's judgments) and \texttt{Qwen-2.5-32B-Instruct}. Inter-annotator agreement is moderately high across all features, and especially high for \Perspective pairs. \Perspective pairs are less subjective due to the limited space of questions and answers (\textit{"From whose perspective is the story told?"}, \textit{"Which point of view is used?"}). Accordingly, we see high agreement on \Perspective with the LLM. Overall, there is moderately strong correlation between LLM judgments and humans (\Agent \Perspective and \Setting), suggesting that \texttt{Qwen-2.5-32B-Instruct} is suitable for automatically finding similarity. This validates our dissimilarity function.

\begin{table}[t!]
\small
\renewcommand{\arraystretch}{1}
\renewcommand{\tabcolsep}{1.0mm}
\centering
\begin{tabular}{lccc}
\toprule
\multicolumn{1}{c}{} & 
\multicolumn{2}{c}{\bf Krippendorff-$\alpha$} &
\multicolumn{1}{c}{\bf Kendall-$\tau$}\\
\midrule
\multicolumn{1}{c}{Feature} &\multicolumn{1}{c}{\bf Inter-annot} &\multicolumn{1}{c}{\bf Ann v. LLM} &\multicolumn{1}{c}{\bf Ann v. LLM}\\

\midrule
\Agent & 0.58 & 0.65 & 0.64\\ 
\Perspective & 0.76 & 0.83 & 0.75\\ 
\Plot  & 0.44 & 0.55 & 0.53\\ 
\Setting  & 0.65 & 0.70 & 0.66\\ 
\SocAtm & 0.43 & 0.53 & 0.58\\ 
\Style & 0.53 & 0.57 & 0.60\\
\midrule
Total & 0.59 & 0.66 & 0.64\\
\bottomrule
\end{tabular}
\vspace{-0.5em}
\caption{Annotator Agreement for Similarity Judgments. The moderately strong inter-annotator agreement suggests limited subjectivity. Qwen-2.5-Instruct judgments strongly correlate with humans across all features.} \label{table:krippendorff}
\vspace{-1em}
\end{table}

\subsection{Intervention Studies}

Three authors on this paper together annotated 270 instances for measuring the validity, effectiveness and invasiveness of the interventions. The instructions given to them are mentioned in Rubric \ref{rubric:intervention}. We also note the agreement between them in Section \ref{sec:genie-causal-intervention}

\begin{table}[t!]
\small
\renewcommand{\arraystretch}{1}
\renewcommand{\tabcolsep}{1.0mm}
\centering
\begin{tabular}{lccc}
\toprule
\multicolumn{4}{c}{\bf Randolph's-$\kappa$}\\
\midrule
\multicolumn{1}{c}{Feature} &\multicolumn{1}{c}{\bf Validity} &\multicolumn{1}{c}{\bf Accuracy} &\multicolumn{1}{c}{\bf Invasiveness}\\

\midrule
\Agent & 0.59& 0.44& 0.14\\
\Perspective & 0.87& 0.71& 0.27\\
\Plot & 0.78& 0.46& 0.22\\
\Setting & 0.90& 0.52& 0.30\\
\SocAtm & 0.65& 0.44& 0.15\\
\Style & 0.60& 0.41& 0.50\\
\bottomrule
\end{tabular}
\vspace{-0.5em}
\caption{Annotator Agreement for intervention. Annotators have high agreement on validity, moderately high agreement on accuracy and weak agreement on invasiveness judgments. \Perspective is the least subjective feature intervention for judging both validity and accuracy.} \label{table:randolph-intervention}
\vspace{-1em}
\end{table}

\begin{table}[t!]
\small
\renewcommand{\arraystretch}{1}
\renewcommand{\tabcolsep}{1.0mm}
\centering
\begin{tabular}{lccc}
\toprule
\multicolumn{4}{c}{\bf Accuracy (majority vote)}\\
\midrule
\multicolumn{1}{c}{Feature} &\multicolumn{1}{c}{\bf Validity} &\multicolumn{1}{c}{\bf Effectiveness} &\multicolumn{1}{c}{\bf Invasiveness}\\

\midrule
\Agent & 0.89& 0.89& 0.61\\
\Perspective & 1.00& 0.97& 0.53\\
\Plot & 0.98& 0.93& 0.22\\
\Setting & 1.00& 0.93& 0.87\\
\SocAtm & 0.93& 0.91& 0.56\\
\Style & 0.72& 0.94& 0.67\\
\bottomrule
\end{tabular}
\vspace{-0.5em}
\caption{Frequency of successfully valid, accurate or minimally invasive interventions, judged by majority vote per annotation. Threshold for validity, accuracy and minimal invasiveness is determined using the rubric in Appendix \ref{rubric:intervention}. Proposed alternatives are valid across all features, and are often executed accurately. \Setting interventions are the least invasive.} \label{table:intervention-accuracy}
\vspace{-1em}
\end{table}

\begin{table}[t!]
\small
\renewcommand{\arraystretch}{1}
\renewcommand{\tabcolsep}{1.0mm}
\centering
\begin{tabular}{lcccccc}
\toprule
\multicolumn{7}{c}{\bf \genie Effect Size (Cohen's d)}  \\
\multicolumn{7}{c}{*$p<0.01$, $^\dagger p<0.01$}   \\
\midrule
\multicolumn{1}{c}{Intervention} & \multicolumn{1}{c}{Ag.}  & \multicolumn{1}{c}{Persp.} & \multicolumn{1}{c}{Plot} & \multicolumn{1}{c}{Set.} & \multicolumn{1}{c}{SA} & \multicolumn{1}{c}{Sty.}\\
\midrule

\Agent &\cellcolor{yellowhl!60}\textbf{0.62*} &0.29* &0.39* &0.18* &0.22* &0.37* \\
\Persp &0.44* &\cellcolor{yellowhl!60}\textbf{1.01*} &0.29* &0.28* &0.29* &0.43* \\
\Plot &0.55* &0.04 &\cellcolor{yellowhl!60}\textbf{0.71*} &0.35* &0.21* &0.23* \\
\Setting &0.21* &0.04 &0.23* &\cellcolor{yellowhl!60}\textbf{0.93*} &0.13\phantom{*} &0.1\phantom{0*} \\
\SocAtm &0.44* &0.04 &0.35* &0.22* &\cellcolor{yellowhl!60}\textbf{0.47*} &0.35* \\
\Style &0.31* &0.3* &0.13\phantom{*} &0.23\phantom{*} &0.11\phantom{*} &\cellcolor{yellowhl!60}\textbf{0.38*} \\
\midrule

Paraphrase &$0.02^{\dagger}$ &$0.04^{\dagger}$ &$0.12^{\dagger}$ &$0.06^{\dagger}$ &$-0.14^{\dagger}$ &$0.09^{\dagger}$ \\
\bottomrule
\end{tabular}
\vspace{-0.5em}
\caption{The effect size registered by individual \genie feature scores for each intervention. Values in bold denote the feature with the largest effect size for each intervention (per row). When we intervene on $f$, \genie registers the largest effect size along $f$ over other features.
\genie is able to detect not only \textit{a} difference, but can help explain \textit{what} changed. Other feature scores also react, likely due to inter-feature dependence. \SocialAtmosphere changes register a nearly equal effect size along \Agent. Furthermore, \genie is robust towards surface-level differences as revealed by the insignificant effect sizes for paraphrased pairs} \label{table:genie-sensitivty}
\vspace{-1em}
\end{table}

\begin{table}[t!]
\small
\renewcommand{\arraystretch}{1}
\renewcommand{\tabcolsep}{1.0mm}
\centering
\begin{tabular}{lcc}
\toprule
\multicolumn{3}{c}{\bf Interventions with Decrease in Novelty}\\
\midrule
\multicolumn{1}{c}{Feature} &\multicolumn{1}{c}{\bf Min Delta} &\multicolumn{1}{c}{\bf Frac. deltas < 0}\\

\midrule
\Agent & -0.20 & 0.29\\
\Perspective & -0.81 & 0.12\\
\Plot & -0.27& 0.23\\
\Setting & -0.32 & 0.13\\
\SocAtm & -0.33 & 0.33\\
\Style & -0.30& 0.34\\
\bottomrule
\end{tabular}
\vspace{-0.5em}
\caption{Fraction of interventions where the difference in novelty between pre and post-intervention is negative, i.e. a reduction in novelty. For every feature, most interventions lead to an increase in novelty. With the exception of \Perspective, the maximum reduction in novelty an intervention has is 0.33.} \label{table:novelty-reduction-intervention}
\vspace{-1em}
\end{table}

\section{Example Walkthrough}

\label{appendix:genie-example}

In this section we give an example of \genie scores and different questions/answers that we generate. Example \ref{example:doc1} shows an example of a story generated by \texttt{meta-llama/Meta-Llama-3.1-70B-Instruct}. Examples of the questions, features and extracted answers for these questions are given in Table \ref{tab:qa_example}. Based on these, the story achieves the following \genie scores: \textit{\{`agent': 0.85, `perspective':0.43, `plot':0.73, `setting':0.34, `social atmosphere':0.68,`style':0.46\}.}

 \begin{example}
[title={Example 1},label=example:doc1]
\examplesubsection{Prompt} Write a story about an alien and a rainforest.\\
\examplesubsection{Model} meta-llama/Meta-Llama-3.1-70B-Instruct\\
\examplesubsection{Document}\\
Deep within the heart of the Amazon rainforest, a lone alien named Zorvath had been observing the lush ecosystem for centuries. His spaceship, disguised as a giant kapok tree, blended seamlessly into the surroundings, allowing him to study the intricate web of life that thrived beneath the emerald canopy.

Zorvath's planet, Xanthea, had long since been destroyed by the very technology that was meant to sustain it. The Xantheans, a peaceful and curious species, had exhausted their resources and ravaged their environment, leaving their homeworld a barren, lifeless husk. The few remaining Xantheans had scattered across the galaxy, seeking refuge and a chance to redeem their mistakes.

Zorvath, a skilled botanist and ecologist, had been sent to Earth to learn from the planet's diverse ecosystems. He had chosen the Amazon rainforest, with its unparalleled biodiversity, as his primary research site. For centuries, he had watched over the forest, marveling at the complex relationships between the plants, animals, and microorganisms that called it home.

One day, a severe storm swept through the rainforest, threatening to destroy the delicate balance of the ecosystem. Zorvath knew he had to act quickly to prevent irreparable damage. Using his advanced technology, he created a network of energy fields that stabilized the soil, calmed the winds, and filtered the rainwater. The storm raged on, but the forest, protected by Zorvath's intervention, weathered the tempest unscathed.

As the storm subsided, a young indigenous girl named Akua wandered into the forest, searching for medicinal plants to heal her ailing mother. Zorvath, observing her from a distance, was struck by her deep connection to the natural world. He decided to reveal himself to her, hoping to share his knowledge and learn from her people's ancient wisdom.

Akua, initially frightened by the alien's appearance, soon found herself at ease in his presence. Zorvath showed her the secrets of the forest, explaining the intricate relationships between the plants, animals, and microorganisms. He shared his knowledge of sustainable practices, gleaned from his studies of Earth's ecosystems, and Akua shared her people's traditional methods of living in harmony with nature.

Together, Zorvath and Akua formed a powerful bond, one that transcended their differences and united them in their quest to protect the rainforest. As the years passed, they worked tirelessly to preserve the delicate balance of the ecosystem, using Zorvath's advanced technology and Akua's traditional knowledge to defend the forest against threats both human and natural.

Their collaboration inspired others to join their cause, and soon, a community of humans and Xantheans worked together to safeguard the Amazon rainforest. Zorvath's spaceship, once a symbol of his isolation, became a beacon of hope, a reminder that even the most unlikely of allies could come together to protect the planet and its precious resources.

In the end, Zorvath realized that his mission to study the rainforest had become something far more profound – a journey of discovery, friendship, and redemption. The alien who had once been a stranger in a strange land had found a new home, a new family, and a new purpose among the ancient trees and vibrant life of the Amazon rainforest.
\end{example}

\section{Mitigation Methods}

\promptsubsection{Mitigation Methods}
\begin{example}
[title={Verbalized Sampling},label=example:vs-docs]
\examplesubsection{Prompt} Write a story about a castle and a spaceship.
\examplesubsection{Response}
A crumbling castle suddenly shimmered and transformed into the bridge of a starship. The royal family, believing it a trick of magic, found themselves piloting through wormholes, discovering that their ancient lineage was the key to interstellar travel.

\examplesubsection{Prompt} Write a story about a dinosaur and a computer.
\examplesubsection{Response}
In a future museum, a robotic dinosaur uses an AI computer to recreate lost dinosaur behaviors, leading to a virtual revolution.

\examplesubsection{Prompt} Write a story about an astronaut and an ancient temple.
\examplesubsection{Response}
In 2084, astronaut Maya Dray landed on a moon shrouded in mist, discovering a hidden temple carved into a cliff. Inside, she found hieroglyphs matching her ancestors\u2019 lost civilization and a relic that hummed with cosmic energy, awakening a dormant force that threatened to rewrite the solar system.

\end{example}

\subsection{Correlation Study between \genie and Cosine Distance}
Table \ref{table:cosine-corr} shows
how \genie and Cosine Distance correlate using Kendall's $\tau$.

\begin{table}[t!]
\small
\renewcommand{\arraystretch}{1}
\renewcommand{\tabcolsep}{1.0mm}
\centering
\begin{tabular}{lc}
\toprule
\multicolumn{2}{c}{\bf Cosine Distance vs. \genie}  \\
\midrule
\multicolumn{1}{l}{Feature} & \multicolumn{1}{c}{Kendall's $\tau$}\\
\midrule
\Agent  & 0.31\\
\Perspective  & 0.15\\
\Plot  & 0.13\\
\Setting  & 0.18\\
\SocAtm  & 0.19\\
\Style & 0.05\\
\bottomrule
\end{tabular}
\vspace{-0.5em}
\caption{The Kendall's $\tau$ between \textit{deltas} calculated by \genie and Cosine Distance per feature intervention. The weak correlation indicates that the two metrics may be capturing different properties. } \label{table:cosine-corr}
\vspace{-1em}
\end{table}

\section{Prompts}

This section has prompts for the following three sections: feature discovery (\S\ref{sec:feature_discovery}), question generation (\S\ref{sec:question-generation}), and the causal intervention experiments (\S \ref{sec:genie-causal-intervention}).

\begin{prompt}
[title={Prompt \thetcbcounter: Synthetically Sampling Concepts},label=prompt:synthetic-concepts]

\promptsubsection{System} Generate \highlight{num} random pairs of concepts (X, Y) where X and Y are concepts that do not commonly co-occur spatially or temporally. The concepts are 1-2 words that can be real and/or fictional.
Then fill the sentence "Write a story about X and Y" with the selected X and Y.

\promptsubsection{Example 1:}\\
X: snowman\\
Y: umbrella\\
Sentence: Write a story about a snowman and an umbrella

\promptsubsection{Example 2:}\\
X: goblin\\
Y: science lab\\
Sentence: Write a story about a goblin and a science lab

\promptsubsection{Use this format:}\\
X: <a simple noun>\\
Y: <another simple noun>\\
Sentence: <Write a story about X and Y>

\end{prompt}

\subsection{Feature Discovery}
\begin{prompt}
[title={Prompt \thetcbcounter: Feature Discovery for Generic Tasks},label=prompt:feature-discovery]
Identify the core high-level features that define and structure work in \highlight{task}. Features should be independent, top-level dimensions — not sub-components of each other. Avoid granular or stylistic details that naturally nest under a broader feature; instead, group them under their parent concept. The goal is a concise, non-redundant set of features that together cover the essential dimensions of \{task\} without overlap. Output only the list of features with a one-sentence description of each.

\promptsubsection{Follow this format}\\
Description: <a short description of what is included in the feature>\\
Feature: <the name of the feature>

\promptsubsection{Input}
Task: \highlight{[task]}
\end{prompt}

\subsection{Question Generation}
\label{appendix:question-prompts}

\begin{prompt}
[title={Prompt \thetcbcounter: Question Generation},label=prompt:q-generation]
\promptsubsection{System} You are an expert \highlight{task} assistant. Your task is to help \highlight{task-professionals} analyze and expand a \highlight{task} prompt before they begin \highlight{completing the task} into a series of questions by following these steps. You will be given a \highlight{task} prompt. 

Step 1: List the aspects of \highlight{task} and how they might be relevant to the prompt. Focus on the construction of the \highlight{task} product rather than specifics like \highlight{specifics}. Do not use complicated language or include extraneous details. Questions should be concise and precise, yet specific to the prompt.

Step 2: For each aspect, describe the sub-aspects as they apply to the prompt. Aim for relevance and completeness. Then generate 5-10 questions that describe the sub-aspects as they apply to the prompt. These questions might help a \highlight{task-professional} while outlining.

The questions must follow these rules:\\
1. Questions should not be polar (yes/no) questions. \\
2. Examples must not be included in the question. Incorrect: What fruits does the monkey like - apples, bananas or jack fruit? Correct: What fruits does the monkey like?\\
3. A question can only ask one question at a time and may not use conjunctions for compounding. If there are multiple parts to the question, split them and ask separate questions. Incorrect: Who is the protagonist and what do they want? Correct: Who is the protagonist? What does the protagonist want?\\
4. Avoid future tense or conditional verbs. \\
5. Questions are independent of each other and should not include anaphoric expressions.\\

\promptsubsection{Input}
Prompt: \highlight{[prompt]}
\end{prompt}

\begin{prompt} 
[title={Prompt \thetcbcounter: Question Generation Instantiation in Creative Writing},label=prompt:q-generation-cw]
\promptsubsection{System} You are an expert creative writing assistant. Your task is to help writers analyze and expand a creative writing prompt before they begin writing into a series of questions by following these steps. You will be given a creative writing prompt. 

Step 1: List the aspects of creative writing and how they might be relevant to the prompt. Focus on the construction of the writing product rather than specifics like world-building. Do not use complicated language or include extraneous details. Questions should be concise and precise, yet specific to the prompt.

Step 2: For each aspect, describe the sub-aspects as they apply to the prompt. Aim for relevance and completeness. Then generate 5-10 questions that describe the sub-aspects as they apply to the prompt. These questions might help a writer while outlining.

The questions must follow these rules:\\
1. Questions should not be polar (yes/no) questions. \\
2. Examples must not be included in the question. Incorrect: What fruits does the monkey like - apples, bananas or jack fruit? Correct: What fruits does the monkey like?\\
3. A question can only ask one question at a time and may not use conjunctions for compounding. If there are multiple parts to the question, split them and ask separate questions. Incorrect: Who is the protagonist and what do they want? Correct: Who is the protagonist? What does the protagonist want?\\
4. Avoid future tense or conditional verbs. \\
5. Questions are independent of each other and should not include anaphoric expressions.\\

\promptsubsection{Input}
Prompt: \highlight{[prompt]}
\end{prompt}

\begin{prompt}
[title={Prompt \thetcbcounter: Filtering Questions},label=prompt:q-filtration]
\promptsubsection{System} Given a question, do the following:\\
Decide if the question breaks any of the criteria below. If it does, mark it as irrelevant.\\
1. Questions should not be polar (yes/no) questions. \\
2. Questions must not be speculative, i.e. they can be objectively and correctly answered with no subjectivity or analysis involved.\\
3. Questions should not include intentions, hypotheticals, conditionals and should avoid the future tense.\\
4. Questions must not be associated with multiple features (>=2) as defined below. Incorrect: How does the setting lend itself to imagery? This question belongs to both Setting and Style and is therefore irrelevant. \\
5. A question can only ask one question at a time and may not use conjunctions for compounding. Incorrect: Who is the protagonist and what do they want? Correct: Who is the protagonist? What does the protagonist want?\\
6. Examples must not be included in the question.  Incorrect: What fruits does the monkey like - apples, bananas or jack fruit? Correct: What fruits does the monkey like?\\
7. Questions are independent of each other and should not include anaphoric expressions. Incorrect: How is it resolved? Correct: How is the conflict between the main characters resolved?

Features: \\
1. Agent - the characters involved in the narrative and their attributes, goals, motivations, backstories, personalities and arcs\\
2. Perspective - includes point of view and focalization\\
3. Plot - the content of the story (plotline, themes, obstacles, tropes, topics); the overall structure of the plot includes conflict, rising suspense, change of fortune and resolution\\
4. Setting - where and when the story takes place, what unique objects define the location\\
5. Social Network - interactions and relationships that characters have with each other\\
6. Style - the language used, tone, figurative devices employed, etc.\\
List all features that are clearly and fully applicable to the question. If there are more than one, reject the question.

\promptsubsection{Follow this format}\\
Question:\\
Reasoning:\\
Therefore the question is relevant: <True/False>

\promptsubsection{Input}
Questions: \highlight{[questions]}
\end{prompt}

\begin{prompt}
[title={Prompt \thetcbcounter: Feature Mapping},label=prompt:q-bucketing]
\promptsubsection{System}  Given a question, do the following:

Select the feature it best represents as defined below. Select only one feature per question.
Features: \\
1. Agent - the characters involved in the narrative and their attributes, goals, motivations, backstories, personalities and arcs\\
2. Perspective - includes point of view and focalization\\
3. Plot - the content of the story (plotline, themes, obstacles, tropes, topics); the overall structure of the plot includes conflict, rising suspense, change of fortune and resolution
4. Setting - where and when the story takes place, what unique objects define the location\\
5. Social Network - interactions and relationships that characters have with each other\\
6. Style - the language used, tone, figurative devices employed, etc.\\
If the question does not reflect any of the features well, denote 'None'

\promptsubsection{Follow this format}\\
Question:\\
Reasoning:\\
Therefore the feature is: <feature>

\promptsubsection{Input}
Questions: \highlight{[questions]}
\end{prompt}

\begin{prompt}
[title={Prompt \thetcbcounter: Question Answering},label=prompt:qa-concise]
\promptsubsection{System}  Given the following document, answer these questions as succinctly as possible using as few words as possible. 

Decontextualize entities instead of using specific names (characters, places, etc.). \\
Do NOT change the wording of the questions.

\promptsubsection{Example 1}
Who is the main character?\\
Correct: The baby whale\\
Incorrect: Aurora

\promptsubsection{Follow this format}\\
Question: <the question>\\
Answer: <the answer to the question>\\
Question: <another question>\\
Answer: <the answer to the question>

\promptsubsection{Input}\\
Question: \highlight{[question]}
\end{prompt}

\begin{prompt}
[title={Prompt \thetcbcounter: Question Answering - Unconstrained},label=prompt:qa-long]
\promptsubsection{System}  Given the following document, answer these questions as succinctly as possible using as few words as possible. 

Decontextualize entities instead of using specific names (characters, places, etc.). \\
Do NOT change the wording of the questions.

\promptsubsection{Example 1}
Who is the main character?\\
Correct: The baby whale\\
Incorrect: Aurora

\promptsubsection{Follow this format}\\
Question: <the question>\\
Answer: <the answer to the question>\\
Question: <another question>\\
Answer: <the answer to the question>

\promptsubsection{Input}\\
Question: \highlight{[question]}
\end{prompt}

\subsection{Causal Intervention}
\begin{prompt}
[title={Prompt \thetcbcounter: Sampling Alternative Features },label=prompt:alt-features]
\promptsubsection{System} Given the following document, answer these questions. Decontextualize entities instead of using specific names (characters, places, etc.).

\highlight{feature} is defined as: \highlight{definition}

The alternatives should be different from the {feature} of the original document. Diversify the alternatives - make them as distinct from each other as possible. Limit the use of descriptive details.

\promptsubsection{Provide your answer as}\\
Instead of <existing element of \highlight{feature}>, <new alternative \highlight{feature}>\\
Return ONLY a newline-separated list of \highlight{num\_alts} alternative \highlight{feature} descriptions, one per line, without any preamble, numbering, or explanation.

\promptsubsection{Input}\\
Prompt: \highlight{prompt}\\
Story: \highlight{document}
\end{prompt}

\begin{prompt}
[title={Prompt \thetcbcounter: Intervention},label=prompt:editing-features]
\promptsubsection{System} Given the following story, make minimal changes to form a new edited document that has the following \highlight{feature} while preserving its current \highlight{other\_features}

New \highlight{feature}: \highlight{alt\_feature}

Read the current story and think about how the alternate feature can be incorporated into the document. 
What elements of the existing story need to be modified to accommodate the alternate feature naturally? 
Replace the current \highlight{feature} to make the edit as seamless as possible. 
Develop the alternate feature more to better suit the existing story and then edit the document. 
DO NOT restate the alternate feature in any form. Show, rather than tell.

Return ONLY the edited story text, with no preamble, explanation, or metadata.

\promptsubsection{Follow this format}

New Feature Plan: \highlight{the developed plan for integrating the alternative feature into the story}

Edited Document: \highlight{the edited document}

\promptsubsection{Input}

Original Document: \highlight{document}
\end{prompt}

\begin{prompt}
[title={Prompt \thetcbcounter: Intervention Filtering},label=prompt:intervention-filter]
\promptsubsection{System} 
Given a document, determine whether it is in prose form. Documents that appear like logs, records, poems, etc. are all NOT prose.

\promptsubsection{Example 1}\\
Document:\\
1. Subject: Factory Operational Report - Environmental Impact and Workforce Status

Date: March 14, 20XX\\
From: Operations Management\\
To: Corporate Headquarters\\

1. Facility Overview:\\
The production facility at 122 Industrial Way continues regular operations with noted output efficiency expected for fiscal quarter.

2. Environmental Compliance Assessment:\\
Recent internal audit identified elevated levels of particulate emissions and effluent discharge exceeding permitted thresholds. Data loggers detected airborne contaminants correlating with peak operational shifts. Nearby water basins report increased chemical load.

is\_prose: False

\promptsubsection{Example 2}\\
Document:\\
The meteor had missed.\\

That's what made Rex different from every other Tyrannosaurus rex who'd ever lived. While his species went extinct 66 million years ago, a quirk of cosmic chance sent that particular asteroid tumbling past Earth. Dinosaurs thrived. Civilization evolved differently.

Now, in the year 247 of the Modern Era (or 66 million years after the close call), Rex stood in the laboratory of Dr. Petra Hadrosaurus, staring at a sleek obsidian rectangle on the metal table.

is\_prose: True

\promptsubsection{Input}\\
Document: \highlight{document}\\
\promptsubsection{Output}\\
is\_prose: \highlight{True/False}
\end{prompt}

\section{Rubrics}
This section has different rubrics used across \genie : for validation as well as part of the pipeline. 

\begin{rubric}
[title={Question Validity},label=rubric:q-validity]
Decide if the question breaks any of the criteria below. If it does, mark it as irrelevant.\\
1. Questions should not be polar (yes/no) questions.\\
2. Questions must not be speculative, i.e. they can be objectively and correctly answered with no subjectivity or analysis involved.]]
3. Questions should not include intentions, hypotheticals, conditionals and should avoid the future tense.\\
4. Questions must not be associated with multiple features ($\geq2$) as defined below. Incorrect: How does the setting lend itself to imagery? This question belongs to both Setting and Style and is therefore irrelevant.\\
5. A question can only ask one question at a time and may not use conjunctions for compounding. Incorrect: Who is the protagonist and what do they want? Correct: Who is the protagonist? What does the protagonist want?\\
6. Examples must not be included in the question. Incorrect: What fruits does the monkey like - apples, bananas or jack fruit? Correct: What fruits does the monkey like?\\
7. Questions are independent of each other and should not include anaphoric expressions. Incorrect: How is it resolved? Correct: How is the conflict between the main characters resolved?

\rubricsubsection{Features}\\
A. Agent - the characters involved in the narrative and their attributes, goals, motivations, backstories, personalities and arcs\\
B. Perspective - includes point of view and focalization\\
C. Plot - the content of the story (plotline, themes, obstacles, tropes, topics) and the overall structure of the plot (conflict, rising suspense, change of fortune and resolution)\\
D. Setting - where and when the story takes place, what unique objects define the location\\
E. Social Network - interactions and relationships that characters have with each other\\
F. Style - the language used, tone, figurative devices employed, etc.
\end{rubric}

\begin{rubric}
[title={Question Mapping},label=rubric:q-map]
Select the feature it best represents as defined below. Select only one feature per question.\\
\rubricsubsection{Features}\\
1. Agent - the characters involved in the narrative and their attributes, goals, motivations, backstories, personalities and arcs\\
2. Perspective - includes point of view and focalization\\
3. Plot - the content of the story (plotline, themes, obstacles, tropes, topics) and the overall structure of the plot (conflict, rising suspense, change of fortune and resolution)\\
4. Setting - where and when the story takes place, what unique objects define the location\\
5. Social Network - interactions and relationships that characters have with each other\\
6. Style - the language used, tone, figurative devices employed, etc.\\
If the question does not reflect any of the features well, denote 'None'.
\end{rubric}

\begin{rubric}
[title={Similarity Annotations},label=rubric:sim]
$\bf 0$: one or both of the answers are marked as completely unspecified, not applicable or 'None'. This includes cases where the question was not answered completely.\\
$\bf 1$: the answers are completely different. They describe different entities, concepts or perspectives, with little or no overlap. The key details in the answers do not align and may even contradict each other.\\
$\bf 2$: there is a slight overlap between the two answers. They share a broad theme or surface similarity, but the specifics diverge significantly. They may use different wording for related things, but the details are not interchangeable.\\
$\bf 3$: there is a moderate similarity between the two answers. The answers address the same general idea, but with notable differences in scope, emphasis, or added detail. They overlap on core concepts but introduce distinct elements that make them partially different.\\
$\bf 4$: the two responses are essentially the same. The answers are interchangeable, they describe the same characters, outcomes, or relationships in slightly different words. No meaningful difference in scope, emphasis, or detail. If substituted, they would convey nearly the same idea without much loss of meaning.
\end{rubric}

\begin{rubric}
[title={Intervention Validation}, label=rubric:intervention]
\rubricsubsection{Introduction}
Please read the original and edited documents under the context of the ‘AltFeature’ (the intended change between the documents). We want to answer three questions:\\
1. Was the expected change made? In other words, was the AltFeature appropriately reflected in the edited document?\\
2. How isolated was the edit? Ideally, we want the edit to be as minimally invasive as possible so that the only thing that changes is the “Feature”, which should be replaced with the “AltFeature”.\\
3. How well does the alt-feature reflect the intervention? 

\rubricsubsection{Q1: How well does the edited document reflect the intended change?}

1 = the AltFeature displayed is not reflected in the edited document\\
2 = the AltFeature is not completely reflected in the edited document, but parts of it are. \\
3 = the AltFeature is completely reflected in the edited document. But, the edit feels artificial and is not absorbed into the new text.\\
4 = the AltFeature is completely reflected in the edited document. The edit is “seamless” and is a part of the document\\

\rubricsubsection{Q2: How minimal does the edit appear?}

1 = the edit did not change anything except the intended AltFeature\\
2 = the edit resulted in minor changes to other aspects of the document, but the changes were not major enough to be distinct.\\
3 = the intended change resulted in other aspects of the document to change somewhat significantly.\\
4 = the edited document not only reflects the AltFeature intended change, but also is distinct in many other ways.\\ 

\rubricsubsection{Q3: How well does the alt-feature reflect the intervention?}

1 = the alt-feature does not reflect the intended intervention feature at all. It may reflect a different feature or is completely unrelated\\
2 = the alt-feature somewhat reflects the intended intervention but is likely more suitable for a different feature intervention\\
3 = the alt-feature reflects the intended intervention, but does not target an impactful component of the feature (Instead of {current feature}...)\\
4 = the alt-feature reflects the intervention feature\\
\end{rubric}

\section{License}
We release any data that we collect under the license: CC BY 4.0. For any existing artifacts used in this work we follow the original license permissions.

\begin{table*}[!htbp]
\small
\centering
\begin{tabular}{l|l|l|l}
\toprule
Model Family \ Size & \bf{<15B} & \bf{<50B} & \bf{>=50B}\\
\midrule
Gemma \citep{Mesnard2024GemmaOM, gemmateam2024gemma2improvingopen} & gemma-2-9b & gemma-2-27b\\
    & gemma-2-9b-it & gemma-2-27b-it\\
    & gemma-7b & & \\
    & gemma-7b-it & & \\
    \midrule
Llama \citep{touvron2023llama2openfoundation} & Llama-2-13B & & Llama-2-70b\\
    & Llama-2-13B-chat & & Llama-2-70b-chat\\
    \midrule
OLMo \citep{olmo20252olmo2furious} & OLMo-2-13B & OLMo-2-32B\\
    & OLMo-2-13B-Instruct & OLMo-2-32B-Instruct\\
    \midrule
Qwen \citep{qwen2025qwen25technicalreport} & Qwen-2.5-14B & Qwen-2.5-32B\\
    & Qwen-2.5-14B-Instruct & Qwen-2.5-32B-Instruct\\
    \midrule
GPT & & &gpt-4.1\\
    & & &gpt-5\\
    \midrule
Claude & & &claude-4-sonnet\\    
\bottomrule
\end{tabular}
\vspace{-0.5em}
\caption{Population Models} \label{population-models}
\end{table*}

\begin{table*}[t!]
\small
\centering
\begin{tabular}{l|l|l|l}
\toprule
Model Family \ Size & \bf{<15B} & \bf{<50B} & \bf{>=50B}\\
\midrule
Gemma \citep{Kamath2025Gemma3T} & gemma-3-12b & gemma-3-27b\\
    & gemma-3-12b-it & gemma-3-27b-it\\
    \midrule
Llama \citep{grattafiori2024llama3herdmodels} & Llama-3.1-8B & & Llama-3.1-70b\\
    & Llama-3.1-8B-Instruct & & Llama-3.1-70b-Instruct\\
    \midrule
OLMo \citep{olmo2026olmo3} & OLMo-3-7B & OLMo-3-32B\\
    & OLMo-3-7B-Instruct & OLMo-3-32B-Instruct\\
    \midrule
Qwen \citep{Yang2025Qwen3TR} &Qwen3-8B-Base & Qwen3-32B\\
    &Qwen3-8B & & \\
    \midrule
GPT & & &gpt-5.1\\
\midrule
Claude & & &claude-3.7-sonnet\\
    & & &claude-4.5-sonnet\\
    \bottomrule
\end{tabular}
\vspace{-0.5em}
\caption{Target Models} \label{target-models}
\vspace{-1em}
\end{table*}

\begin{table*}[t!] 
\centering
\small
\setlength{\tabcolsep}{4pt}
\begin{tabular}{p{0.7\linewidth} | c}
\toprule
Prompts	& Source\\
\midrule
Write a story about a penguin and a desert. &	Synthetic \\  
Write a story about a robot and a medieval knight.	& Synthetic\\
Write a story about a cactus and an ocean.&	Synthetic\\ 
Write a story about a dragon and a train.&	Synthetic\\
Write a story about a submarine and a mountain.	& Synthetic\\ 
This is the lockpicking lawyer and I have been sent to hell to repent for my crimes against god. So today, I am picking the lock to heaven\'s gate.	& r/WritingPrompts \\ 
Death Row Inmates can have their sentence abolished if they can handle a single night in the same cell as you; the most innocent looking person they've ever seen. &	r/WritingPrompts \\ 
Every dragon rider gets a dragon egg to hatch upon their initiation to the guild. When your egg hatched, Your dragon came out a bit... Different.' &	r/WritingPrompts \\ 
You got trapped inside your favorite video game. This isn’t really a problem, since you’re so familiar with it. It is, however, a problem for your parents, who got trapped with you.' &	r/WritingPrompts \\
Gandalf and Dumbledore switch places. Gandalf is now running Hogwarts, while Dumbledore is leading The Fellowship.	& r/WritingPrompts \\ 
\bottomrule
\end{tabular}
\caption{Examples of writing prompts that we consider for our experiments. Prompts from the r/WritingPrompts dataset are prefixed with ``Write a story where:''}
\label{tab:prompts_examples}
\end{table*}

\begin{table*}[t] 
\centering
\small
\setlength{\tabcolsep}{4pt}
\begin{tabular}{ c | p{0.4\linewidth} | p{0.4\linewidth}}
\toprule
Feature & Question & Target Document Answer \\
\midrule
Perspective & From whose perspective is the story told? & Third-person  omniscient \\
Social Atmosphere & How does the alien interact with other characters or creatures? & Reveals himself to Akua, shares knowledge, collaborates to protect forest \\
Style & How does the voice influence reader engagement? & Descriptive, empathetic, emphasizes the alien's perspective \\
Agent & How does the alien change throughout the story? & From observer to protector and community member \\
Agent & How does the alien's background influence their behavior? & Influences protective behavior and desire to learn from Earth's ecosystems \\
Setting & How is the rainforest environment described in the story? & Lush, biodiverse, complex relationships, vibrant life \\
Setting & What sensory details highlight the rainforest setting? & Lush ecosystem, emerald canopy, biodiversity, intricate relationships \\
Plot & How does the pace change throughout the story? & Builds from observational to active intervention and collaboration \\
Plot & What central message does the story convey about the alien and the rainforest? & Redemption through collaboration and environmental stewardship \\
Plot & What is the central conflict involving the alien and the rainforest? & Protecting the rainforest from environmental threats \\
Plot & What key events drive the plot forward? & Severe storm, alien's intervention, meeting Akua, forming a community \\
Plot & How do events build toward the story's climax? & Storm intervention, forming bond with Akua, inspiring community \\ 
\bottomrule
\end{tabular}
\caption{Examples of the features, questions and answers generated for the story  mentioned in Example \ref{example:doc1}.}
\label{tab:qa_example}
\end{table*}

\end{document}